\documentclass[11pt]{article}

\usepackage[preprint]{acl}

\usepackage{times}
\usepackage{latexsym}

\usepackage[T1]{fontenc}

\usepackage[utf8]{inputenc}

\usepackage{microtype}

\usepackage{inconsolata}

\usepackage{graphicx}
\usepackage{subcaption}
\usepackage{booktabs}
\usepackage{amsmath}
\usepackage{amssymb}
\usepackage{xcolor}
\usepackage{colortbl}
\usepackage[most]{tcolorbox}
\usepackage{placeins}
\definecolor{ourscolor}{RGB}{207,226,243}

\usepackage{enumitem}

\title{\texttt{SLEA}-RL~\raisebox{-0.1em}{\includegraphics[height=1.1em]{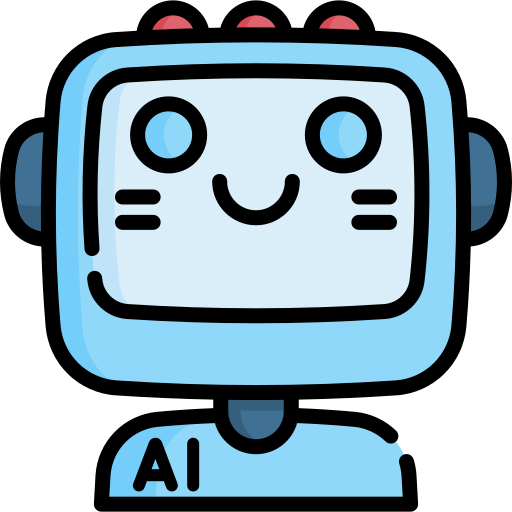}}\raisebox{-0.1em}{\includegraphics[height=1.1em]{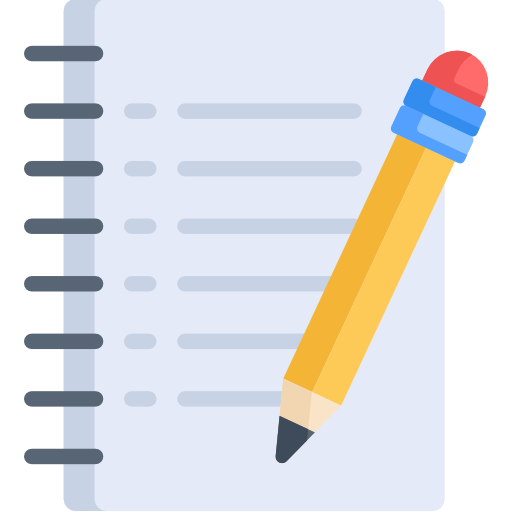}}: Step-Level Experience Augmented Reinforcement Learning for Multi-Turn Agentic Training}

\author{Prince Zizhuang Wang\\
  Carnegie Mellon University \\
  \texttt{princewang@cmu.edu} \\\And
  Shuli Jiang \\
  Carnegie Mellon University\\
  \texttt{shulij@andrew.cmu.edu} \\}

\usepackage{fancyhdr}
\fancypagestyle{firstpage}{%
  \fancyhf{}%
  \renewcommand{\headrulewidth}{0pt}%
  \fancyfoot[L]{\footnotesize\textit{Preprint.}}%
}

\begin{document}
\thispagestyle{firstpage}
\maketitle
\begin{abstract}
    Large Language Model (LLM) agents have shown strong results on multi-turn tool-use tasks, yet they operate in isolation during training, failing to leverage experiences accumulated across episodes. Existing experience-augmented methods address this by organizing trajectories into retrievable libraries, but they retrieve experiences only once based on the initial task description and hold them constant throughout the episode. In multi-turn settings where observations change at every step, this static retrieval becomes increasingly mismatched as episodes progress. We propose \texttt{SLEA}-RL (Step-Level Experience-Augmented Reinforcement Learning), a framework that retrieves relevant experiences at each decision step conditioned on the current observation. \texttt{SLEA}-RL operates through three components: (i) step-level observation clustering that groups structurally equivalent environmental states for efficient cluster-indexed retrieval; (ii) a self-evolving experience library that distills successful strategies and failure patterns through score-based admission and rate-limited extraction; and (iii) policy optimization with step-level credit assignment for fine-grained advantage estimation across multi-turn episodes. The experience library evolves alongside the policy through semantic analysis rather than gradient updates. Experiments on long-horizon multi-turn agent benchmarks demonstrate that \texttt{SLEA}-RL achieves superior performance compared to various reinforcement learning baselines. Our code is available at \url{https://github.com/kingofspace0wzz/slea-rl/}.
\end{abstract}
    
    \section{Introduction}

    \begin{figure}[t]
    \centering
    \includegraphics[width=0.85\columnwidth]{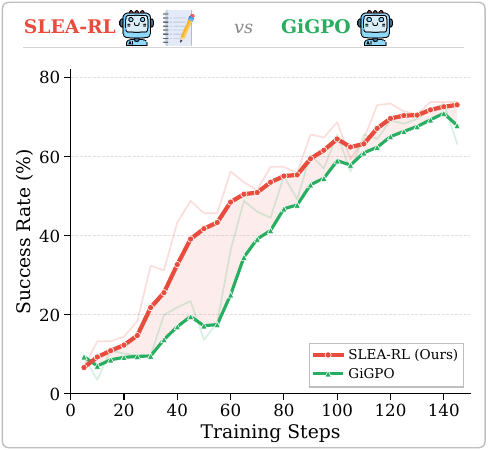}
    \caption{Validation success rate on WebShop (Qwen2.5-1.5B-Instruct). \texttt{SLEA}-RL achieves faster convergence and higher asymptotic performance compared to GiGPO and GRPO. Training curves are included in Appendix~\ref{sec:appendix_add_res}.}
    \label{fig:motivating}
    \vspace{-2mm}
    \end{figure}

    Large language model (LLM) agents~\citep{yao2022react, shinn2023reflexion} have demonstrated remarkable capabilities across complex tasks such as web navigation~\citep{yao2022webshop}, embodied planning~\citep{ALFWorld20}, and tool-integrated reasoning~\citep{jin2025searchr1}. Reinforcement learning (RL) has become a key paradigm for training these agents, with group-based algorithms such as GRPO~\citep{guo2025deepseek} and Dr. GRPO~\citep{liu2025understanding} proving especially effective. Despite these advances, each task execution remains largely episodic: current LLM agents operate in isolation, unable to learn from past successes or failures across episodes, which significantly hinders their evolution.

    Multi-turn tool-use tasks---such as ALFWorld~\citep{ALFWorld20}, WebShop~\citep{yao2022webshop}, and interactive web search---present a fundamentally different challenge than single-turn reasoning~\citep{achiam2023gpt, hui2024qwen2, team2023gemini, liu2024deepseek}. Unlike static settings where the context remains fixed throughout generation, multi-turn environments change at every step: each action reveals a new room configuration, each click surfaces new listings. The agent must continuously adapt its reasoning to an ever-changing world state across dozens of sequential decisions.
    Standard RL approaches~\citep{schulman2017proximal, ahmadian2024back, guo2025deepseek, zheng2025gspo, feng2025group} face three compounding challenges in this setting. First, each training rollout acts in isolation: episodes begin from scratch with no memory of prior interactions, causing the agent to repeat mistakes and discard successful strategies across thousands of rollouts. Second, most multi-turn environments provide only \emph{sparse, outcome-level rewards}, leaving dozens of intermediate steps without learning signal and making credit assignment difficult. Third, RL training proceeds purely through \emph{weight updates}, discarding the structured, interpretable knowledge that accumulates across episodes in a form that could directly guide future behavior.

    Recent work on experience-augmented agents~\citep{xia2026skillrl, zhang2026memrl, flex} addresses the first limitation by organizing successful trajectories into retrievable libraries. However, these methods perform \emph{task-level} retrieval: a fixed set of experiences is fetched once based on the initial task description and held constant for the entire episode. In multi-turn settings, this provides increasingly stale context as the environment state diverges from the original task prompt. Hence, these limitations raise a core question:
    \begin{tcolorbox}[colback=ourscolor, colframe=black!80, boxrule=1.5pt, arc=2pt, left=6pt, right=6pt, top=4pt, bottom=4pt]
    \textit{Can LLM agents learn to leverage step-specific accumulated experience throughout an episode, matching retrieved strategies to the current observation at every decision point?}
    \end{tcolorbox}
    The key insight is that relevant accumulated experience is \emph{state-dependent}: the strategies useful when a WebShop agent first navigates to a product category differ fundamentally from those needed when comparing two specific items at checkout twenty steps later. This state-dependence connects naturally to step-level credit assignment:
    
    \begin{tcolorbox}[colback=ourscolor, colframe=black!80, boxrule=1.5pt, arc=2pt, left=6pt, right=6pt, top=4pt, bottom=4pt]
        \textit{How does agent determine which intermediate actions causally contributed to the final outcome particularly when episodes span dozens of steps?}
        \end{tcolorbox}

    Just as not all intermediate actions deserve equal credit for the episode outcome, not all steps benefit equally from the same retrieved experiences. Step-level retrieval addresses both dimensions simultaneously: by conditioning retrieved experiences on the current observation, agents receive targeted support at exactly the steps where it matters, improving the quality of the exploratory actions that receive fine-grained credit.

    To this end, we propose \textbf{\texttt{SLEA}-RL \textit{(Step-Level Experience-Augmented Reinforcement Learning)}}, a multi-turn RL framework that integrates step-level experience retrieval directly into the training loop. At each decision step, \texttt{SLEA}-RL retrieves relevant experiences from a structured experience library conditioned on the current observation, injecting them via format-preserving augmentation. To enable efficient retrieval at scale, \texttt{SLEA}-RL clusters structurally equivalent observations into equivalence classes, so that experiences proven effective at one state immediately benefit all similar states across trajectories. The experience library self-evolves through quality-controlled semantic analysis of the best and worst trajectories each epoch, with score-based admission and rate-limited extraction maintaining quality under continuous update pressure. Crucially, experiences are used during both training and inference, ensuring that the policy learns to effectively leverage retrieved experiences rather than treating them as a test-time artifact.
    Our main contributions are:

    \begin{itemize}[itemsep=0mm, topsep=0mm]
        \item \textbf{Step-level experience retrieval with observation clustering.} Unlike existing methods that retrieve experiences once per task, \texttt{SLEA}-RL retrieves at each decision step conditioned on the current observation, and clusters structurally similar observations for efficient retrieval and automatic cross-trajectory generalization.
        \item \textbf{Quality-controlled self-evolving experience library.} We introduce score-based admission with rate-limited extraction that prevents degradation under continuous update pressure, enabling the experience library to co-evolve with the policy through semantic analysis rather than gradient updates.
        \item \textbf{Empirical validation on long-horizon benchmarks.} \texttt{SLEA}-RL achieves strong results on ALFWorld, WebShop, and seven search-augmented QA tasks, outperforming many standard RL and experience-augmented baselines.
    \end{itemize}
    
    \section{Preliminaries}
    
    \begin{figure*}[htbp]
        \centering
        \includegraphics[width=\textwidth]{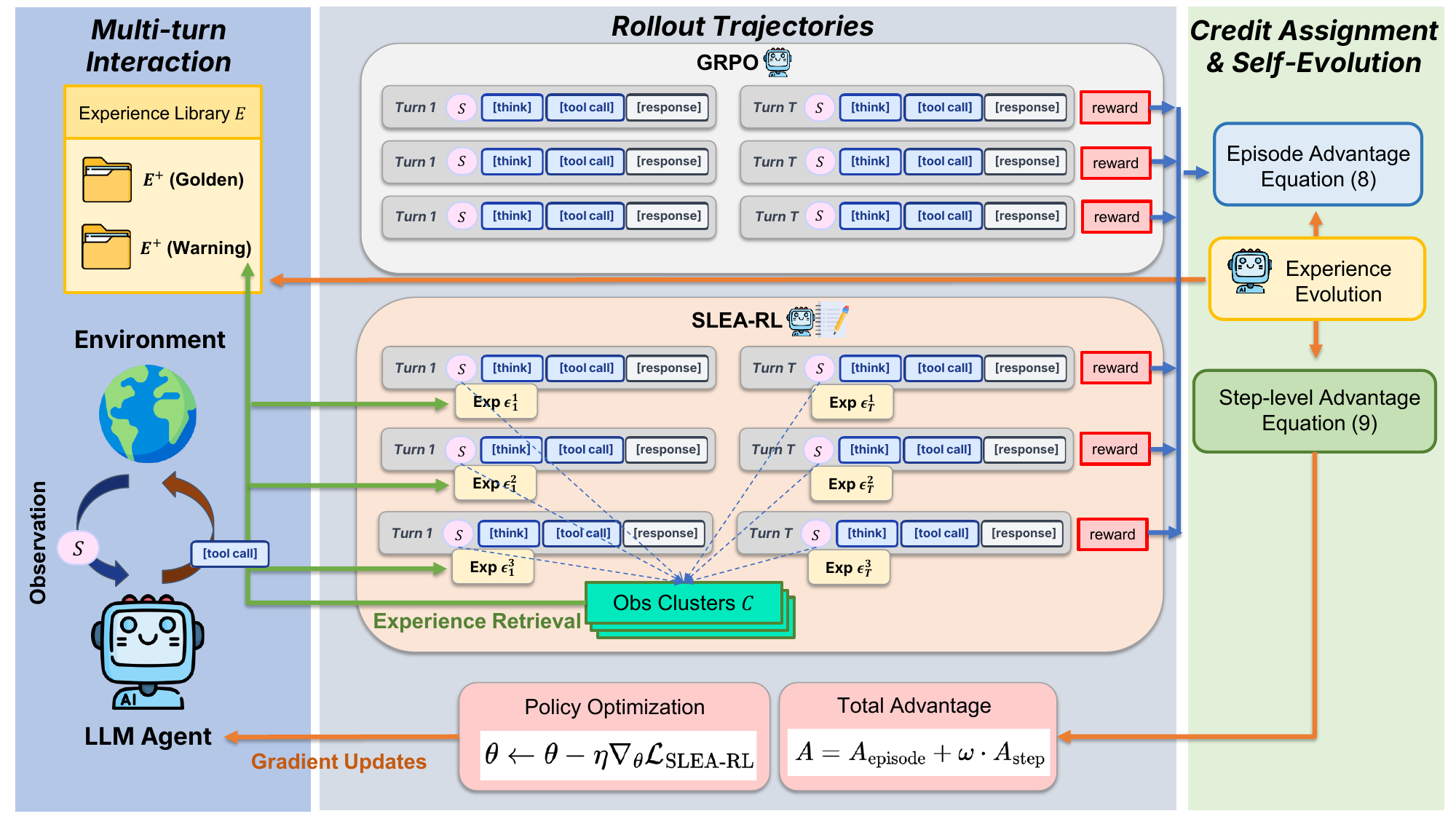}

        \caption{Overview of \texttt{SLEA}-RL compared to standard GRPO. \textbf{Left}: the LLM agent interacts with the environment in a multi-turn loop; an experience library $E$ (comprising strategy zone $E^+$ and warning zone $E^-$) provides step-level guidance. \textbf{Center}: GRPO (top, gray) samples $G$ trajectories without experience augmentation, while \texttt{SLEA}-RL (bottom, orange) retrieves experiences $\varepsilon_{t}^{(i)}$ at each step via cluster-indexed lookup over observation clusters $\mathcal{C}$. }
        \label{fig:pipeline}
    \end{figure*}

    \subsection{LLM Agent Framework}
    
    We consider an LLM agent operating in an interactive environment $\mathcal{E}$. At each timestep $t$, the agent observes a state $o_t \in \mathcal{O}$, selects an action $a_t \in \mathcal{A}$, and receives a reward $r_t$ and next observation $o_{t+1}$. A trajectory $\tau = (o_0, a_0, r_0, \ldots, o_T, a_T, r_T)$ captures one episode of interaction. Tasks are specified by natural language descriptions $d$. An LLM-based agent parameterized by $\theta$ implements a policy $\pi_\theta(a_t | o_{\leq t}, d, c)$ where $c$ represents additional context (e.g., retrieved experiences). Our goal is to learn a policy that maximizes expected return $\max_\theta \mathbb{E}_{\tau \sim \pi_\theta}\left[\sum_{t=0}^T \gamma^t r_t\right]$ subject to context length constraints $|c| \leq L_{\max}$.
    In many environments, the agent receives only a sparse terminal reward $R(\tau) \in \{0, 1\}$ indicating task success or failure, with intermediate rewards $r_t = 0$ for all $t < T$. In this setting, the discounted return from step $t$ reduces to $\hat{R}_t = \gamma^{T-t} R(\tau)$. Our framework accommodates both sparse and dense reward settings; we use $r_t$ in full generality and specialize where needed. This sparse reward structure makes step-level credit assignment particularly challenging, as the agent must determine which intermediate actions causally contributed to the final outcome across episodes spanning dozens of steps.
    
    \subsection{Group Relative Policy Optimization}
    
    Group Relative Policy Optimization (GRPO)~\citep{guo2025deepseek} is a reinforcement learning method that avoids training a critic by using intra-group relative rewards to optimize the policy. For each query $x$, the model samples $G$ responses $\{y^{(1)}, \ldots, y^{(G)}\}$, which are scored to obtain rewards $\{R^{(1)}, \ldots, R^{(G)}\}$, where $i \in [G]$ indexes the trajectory within the sampled group. GRPO computes normalized advantages and updates the policy with a PPO-style clipped objective~\citep{schulman2017proximal}:
    \begin{align*}
    \label{eq:grpo}
    &\mathcal{J}_{\text{GRPO}}(\theta) = \mathbb{E}_{x, \{y^{(i)}\}} \bigg[\frac{1}{G} \sum_{i=1}^{G} \min \Big(\rho^{(i)} A^{(i)},\nonumber\\
    &\quad \text{clip}(\rho^{(i)}, 1\!-\!\epsilon, 1\!+\!\epsilon) A^{(i)} \Big) - \beta D_{\text{KL}}(\pi_\theta \| \pi_{\text{ref}})\bigg],
    \end{align*}
    where $\rho^{(i)} = \frac{\pi_\theta(y^{(i)} | x)}{\pi_{\text{old}}(y^{(i)} | x)}$ is the importance ratio and $A^{(i)} = \frac{R^{(i)} - \text{mean}(\{R^{(j)}\}_{j=1}^G)}{\text{std}(\{R^{(j)}\}_{j=1}^G)}$ is the normalized advantage, with $\epsilon$ and $\beta$ as hyperparameters, $\pi_{\text{old}}$ the policy before the current update, and $\pi_{\text{ref}}$ the reference policy for KL regularization. This formulation eliminates value network training while providing automatic baseline subtraction and variance normalization. A critical property for time efficiency is that sampling $G$ trajectories in parallel using modern inference engines incurs wall-clock time similar to sampling a single trajectory, making group-based methods highly practical.

    \section{\texttt{SLEA}-RL~\raisebox{-0.1em}{\includegraphics[height=1.1em]{ai.png}}\raisebox{-0.1em}{\includegraphics[height=1.1em]{notes.png}}}

    We propose \texttt{SLEA}-RL, a multi-turn reinforcement learning framework that integrates step-level experience retrieval into agent training. The framework contains three interleaved components:
    (i) Self-evolving experience library, which distills reusable strategies and warnings from successful and failed trajectories through quality-controlled semantic analysis.
    (ii) Step-level observation clustering with cluster-indexed retrieval, enabling efficient retrieval of relevant past experiences.
    (iii) Experience-augmented rollouts with policy optimization, where retrieved experiences are incorporated into the agent context during training to guide decision-making.

    \subsection{Experience Library}
    \label{sec:experience-library}

    The experience library $E = E^+ \cup E^-$ is a structured repository of textual experiences accumulated during training. Each entry $e \in E$ is represented as a tuple $e=(s_e, l_e, z_e)$, where $s_e$ is a natural-language description of a strategy or warning, $l_e \in \{\text{principle}, \text{pattern}, \text{example}\}$ denotes its abstraction level, and $z_e \in \mathbb{R}$ is a quality score derived from the reward of the trajectory from which the experience is extracted.
    The library is divided into two zones: a \emph{strategy zone} $E^+$, which stores successful strategies organized hierarchically into principles, reasoning patterns, and concrete examples; and a \emph{warning zone} $E^-$, which records failure knowledge such as common mistakes and recurring failure patterns. To control growth, the library capacity is limited to $C$ entries per level in each zone.

    
    We use $\varepsilon_t \subset E$ to denote the subset of experiences retrieved at step $t$ for a given observation. The library evolves through semantic analysis rather than gradient updates (Section~\ref{sec:self-evolving}), providing interpretable knowledge structures that can be inspected and transferred across agents.

    \subsection{Step-level Observation Clustering}
    \label{sec:obs-clustering}
    Next, we introduce the observation clustering mechanism to enable efficient experience retrieval and consistent step-level credit assignment. In multi-turn environments, many observations are semantically similar; clustering allows such states to share both advantages and retrieved experiences. Prior work groups similar observations for advantage normalization~\citep{feng2025group}. \texttt{SLEA}-RL extends this idea by using the same structure to index reusable experiences.

    We maintain a cluster index $\mathcal{C} = \{c_1,\ldots,c_M\}$, where each cluster $c_i$ is represented by a prototype observation $\mathrm{rep}(c_i)$ and stores associated experience pools $E^+_{c_i} \subseteq E^+$ and $E^-_{c_i} \subseteq E^-$. Given a new observation $o_t$, we assign it to an existing cluster or create a new one:
    \begin{align*}
    c(o_t)=
    \begin{cases}
    c_i & \text{if } \exists\, c_i:\ \mathrm{sim}(o_t,\mathrm{rep}(c_i)) \ge \delta \\
    \text{new cluster} & \text{otherwise.}
    \end{cases}
    \end{align*}
    where $\text{sim}(\cdot, \cdot)$ is a sequence similarity function and $\delta$ is the similarity threshold.
    
    If the assigned cluster already contains experiences, retrieval is performed directly from its pools:
    \begin{equation}
    \label{eq:cluster-retrieval}
    \varepsilon_t =
    \mathrm{TopK}\!\left(E^+_{c(o_t)},k^+\right)
    \cup
    \mathrm{TopK}\!\left(E^-_{c(o_t)},k^-\right),
    \end{equation}
    where $k^+$ and $k^-$ control the number of retrieved strategies and warnings. This cluster-indexed retrieval avoids scanning the full library while ensuring both positive and cautionary guidance.
    
    If the cluster has no associated experiences, the method falls back to library-wide semantic retrieval:
    \begin{equation}
    \label{eq:fallback-retrieval}
    \varepsilon_t =
    \mathrm{TopK}\!\left(\{e \in E : \mathrm{sim}(o_t,e) > \delta\},k\right),
    \end{equation}
    and the retrieved experiences are linked to the cluster for future reuse.
    
    As training progresses, most observations match existing clusters, making retrieval increasingly efficient and targeted. During library evolution (Section~\ref{sec:self-evolving}), newly admitted experiences are associated with relevant clusters, while score-based admission replaces lower-quality entries, allowing cluster pools to improve alongside the policy.

    \subsection{Experience-Augmented Rollout}

    Building the observation clustering, \texttt{SLEA}-RL performs \emph{step-level experience retrieval} during rollout. At step $t$ in a trajectory, the agent observes $o_t$ and retrieves relevant experiences $\varepsilon_t$ using the cluster-indexed mechanism (Eq.~\ref{eq:cluster-retrieval}--\ref{eq:fallback-retrieval}). To avoid injecting low-quality experiences early in training, retrieval is activated only after a warmup of $W$ epochs and when the library size exceeds $|E| > C_{\min}$.
    
    When retrieval is enabled, the policy conditions on both the current observation and retrieved experiences:
    \begin{equation}
    a_t \sim \pi_\theta(\cdot \mid \text{augment}(o_t, \varepsilon_t)),
    \end{equation}
    where experiences are inserted into the prompt via format-preserving augmentation (Appendix~\ref{sec:format-preserving}).
    
    The key distinction from prior work~\citep{rlep, flex} is that experience retrieval occurs \emph{at every step} based on the current observation, rather than once from the initial task description. This design allows experience guidance to adapt to evolving agent states and ensures that the model learns to utilize retrieved experiences during training under the same conditions used at inference.

    \subsection{Self-Evolving Experience Library}
    \label{sec:self-evolving}

    The experience library evolves through \emph{quality-controlled semantic analysis} rather than gradient updates. Naively extracting experiences from all trajectories would generate hundreds of candidates per step, causing eviction churn and unstable retrieval. \texttt{SLEA}-RL addresses this with three controls: outcome partitioning, selective extraction, and score-based admission.
    
    \paragraph{Outcome partitioning.}
    After rollout, trajectories are split by reward. Let $r_j = R(\tau_j)$ denote the reward of trajectory $\tau_j$. Using a threshold $\eta$ (typically the batch median), we form
    \begin{align*}
    \mathcal{T}^+ = \{\tau_j : r_j > \eta\}, \quad
    \mathcal{T}^- = \{\tau_j : r_j \le \eta\},
    \end{align*}
    where successes provide strategy signals and failures reveal pitfalls.
    
    \paragraph{Selective extraction.}
    Only the top-$K_{\text{traj}}$ trajectories from $\mathcal{T}^+$ and bottom-$K_{\text{traj}}$ from $\mathcal{T}^-$ are analyzed for experience extraction. An auxiliary LLM summarizes reusable strategies (from successes) or warnings (from failures). Each extracted experience inherits the episode reward of its source trajectory as its quality score $z_e = R(\tau)$. To limit growth, each evolution step generates at most $K_{\text{strat}}$ strategies and $K_{\text{warn}}$ warnings.

    \paragraph{Score-based admission.}
    Each library zone has capacity $C$. A candidate experience is admitted only if its quality score $z_e$ exceeds the lowest-scoring entry in that zone; otherwise it is discarded. As the policy improves, higher-reward trajectories yield higher-quality experiences that replace older ones, keeping the library aligned with the agent's evolving capabilities. Accepted experiences are inserted into $E^+$ or $E^-$ and associated with relevant observation clusters.

    \subsection{Policy Optimization}

    \texttt{SLEA}-RL combines episode-level and step-level credit assignment. For each task, we sample $G$ trajectories with experience-augmented rollout and compute episode-level advantages following GRPO.
    To provide fine-grained credit, we introduce \emph{step-level advantages} based on observation clusters (Section~\ref{sec:obs-clustering}). For each cluster, we group steps with similar observations across trajectories and normalize their discounted returns within the group. This assigns different credit to actions taken under similar states but leading to different outcomes.
    The final advantage is a weighted combination:
    \begin{equation}
    \hat{A} = A_{\text{episode}} + w \cdot A_{\text{step}},
    \end{equation}
    where $w$ controls the contribution of step-level credit. The policy is then optimized using the standard clipped GRPO objective with $\hat{A}$.
    Importantly, \texttt{SLEA}-RL requires no additional loss terms or auxiliary passes: experience-guided learning arises directly from rollout, where retrieved experiences influence actions and are reinforced through their impact on returns.

    Step-level advantages provide fine-grained credit at individual decision points. For each timestep $t$ across all $G$ trajectories, we assign the observation $o_t^{(i)}$ to its nearest cluster in the embedding space (Section~\ref{sec:obs-clustering}). Steps from different trajectories that map to the \emph{same} cluster $c$ form a step-level group: these are decisions made under similar states but potentially leading to different outcomes. Formally, we collect all (action, discounted return) pairs from steps assigned to $c$ into $\mathcal{S}(c) = \{(a_t^{(i)}, \hat{R}_t^{(i)}) \mid c(o_t^{(i)}) = c\}$, where $\hat{R}_t^{(i)} = \sum_{k=t}^{T} \gamma^{k-t} r_k^{(i)}$ is the discounted return from step $t$, computed from per-step environment rewards $r_k^{(i)}$ (distinct from the episode-level reward $R^{(i)} = R(\tau^{(i)})$). Within each such group, we normalize the discounted returns to obtain per-step advantages:
    \begin{equation}
    \resizebox{0.88\columnwidth}{!}{$\displaystyle
    A_{\text{step}}(a_t^{(i)}) = \frac{\hat{R}_t^{(i)} - \text{mean}\!\left(\{\hat{R}_t^{(j)} \mid (a_t^{(j)}, \hat{R}_t^{(j)}) \in \mathcal{S}(c)\}\right)}{\text{std}\!\left(\{\hat{R}_t^{(j)} \mid (a_t^{(j)}, \hat{R}_t^{(j)}) \in \mathcal{S}(c)\}\right)}
    $}
    \end{equation}
    Actions taken at the same observation cluster but leading to different outcomes receive different credit, enabling the policy to distinguish effective from ineffective decisions at each state. The combined advantage for each action $a_t^{(i)}$ is:
    \begin{equation}
    \hat{A}(a_t^{(i)}) = A_{\text{episode}}(\tau^{(i)}) + w \cdot A_{\text{step}}(a_t^{(i)})
    \end{equation}
    where $w$ controls the relative weight of step-level credit assignment. The policy is updated using the clipped surrogate objective extended with the multi-level advantage:
    \begin{align}
    \label{eq:slearl-obj}
    \mathcal{J}(\theta) &= \mathbb{E}_{d, \{\tau^{(i)}\}} \bigg[\frac{1}{G} \sum_{i=1}^{G} \frac{1}{T_i} \sum_{t=1}^{T_i} \nonumber\\
    &\;\; \min \!\Big(\rho_t^{(i)} \hat{A}_t^{(i)},\, \text{clip}(\rho_t^{(i)}, 1\!-\!\epsilon, 1\!+\!\epsilon)\, \hat{A}_t^{(i)} \Big) \nonumber\\
    &\;\; - \beta\, D_{\text{KL}}(\pi_\theta \| \pi_{\text{ref}})\bigg],
    \end{align}
    where $\rho_t^{(i)} = \frac{\pi_\theta(a_t^{(i)} | o_{\leq t}^{(i)}, d, \varepsilon_t)}{\pi_{\text{old}}(a_t^{(i)} | o_{\leq t}^{(i)}, d, \varepsilon_t)}$ is the per-step importance ratio conditioned on the experience-augmented context.
    \section{Experiments}

    We present empirical evaluations of \texttt{SLEA}-RL across a variety of agentic tasks. Our experiments aim to answer the following questions: \textbf{(1)}~How does \texttt{SLEA}-RL compare to state-of-the-art RL and experience-augmented baselines? \textbf{(2)}~What is the contribution of each component (step-level retrieval, observation clustering, self-evolving library)? \textbf{(3)}~How does step-level retrieval compare to task-level retrieval? \textbf{(4)}~How does the experience library scale with training?

    \subsection{Experimental Setup}

    \textbf{Benchmarks.}\quad
    We evaluate on two multi-turn agent benchmarks, \textit{ALFWorld}~\citep{ALFWorld20} and \textit{WebShop}~\citep{yao2022webshop}, as well as search-augmented QA tasks covering both single-hop and multi-hop settings. See Appendix A for details.

    \textbf{Baselines.}\quad
    For ALFWorld and WebShop, we compare against: (a)~\textit{closed-source LLMs} (GPT-4o, Gemini-2.5-Pro); (b)~\textit{prompt-based and memory methods} that guide behavior via in-context reasoning or external memory without parameter updates (ReAct, Reflexion, Mem0, ExpeL, MemP, SimpleMem); (c)~\textit{RL methods} with trajectory-level or group-based advantage estimation (RLOO, GRPO, GiGPO, R$^3$L, IGPO); and (d)~\textit{experience-augmented RL} that integrates persistent memory into the RL loop (MemRL, Mem0+GRPO, SimpleMem+GRPO, SkillRL). For search-augmented QA, we compare against R1-Instruct, Search-R1, ZeroSearch, StepSearch, EvolveR, and RL baselines PPO, Reinforce++, GSPO, and GiGPO.

    \textbf{Implementation details.}\quad
    We use Qwen2.5-7B-Instruct as the primary base model, and additionally evaluate Qwen2.5-1.5B-Instruct to study scaling. RL training uses a learning rate of $1\!\times\!10^{-6}$, batch size 16, group size $G{=}8$, and 4 gradient accumulation steps, with clipping $\epsilon{=}0.2$ and KL coefficient $\beta{=}0.01$. The step-level advantage weight is $w{=}1.0$.

    For the experience library, we set the similarity threshold $\delta{=}0.85$, warmup $W{=}5$ epochs, and retrieve top-$2$ strategies and top-$1$ warnings per step. Experience extraction and semantic analysis are performed by the same base model used for rollout, requiring no additional model deployment. Per-step retrieval adds minimal overhead: cluster-indexed lookup uses string similarity ($O(|\mathcal{C}|)$ comparisons) rather than neural embeddings, and the retrieved text is prepended to the existing prompt without additional forward passes. The main computational addition is the auxiliary LLM call for experience extraction, which runs once per training batch. For search-augmented QA, we follow the same setup as Search-R1, using E5 as the retriever with group size $G{=}5$. Full details are provided in Appendix~\ref{sec:hyperparams}.

    \begin{figure*}[t]
        \centering
        \begin{subfigure}[t]{0.49\textwidth}
            \centering
            \includegraphics[width=\textwidth]{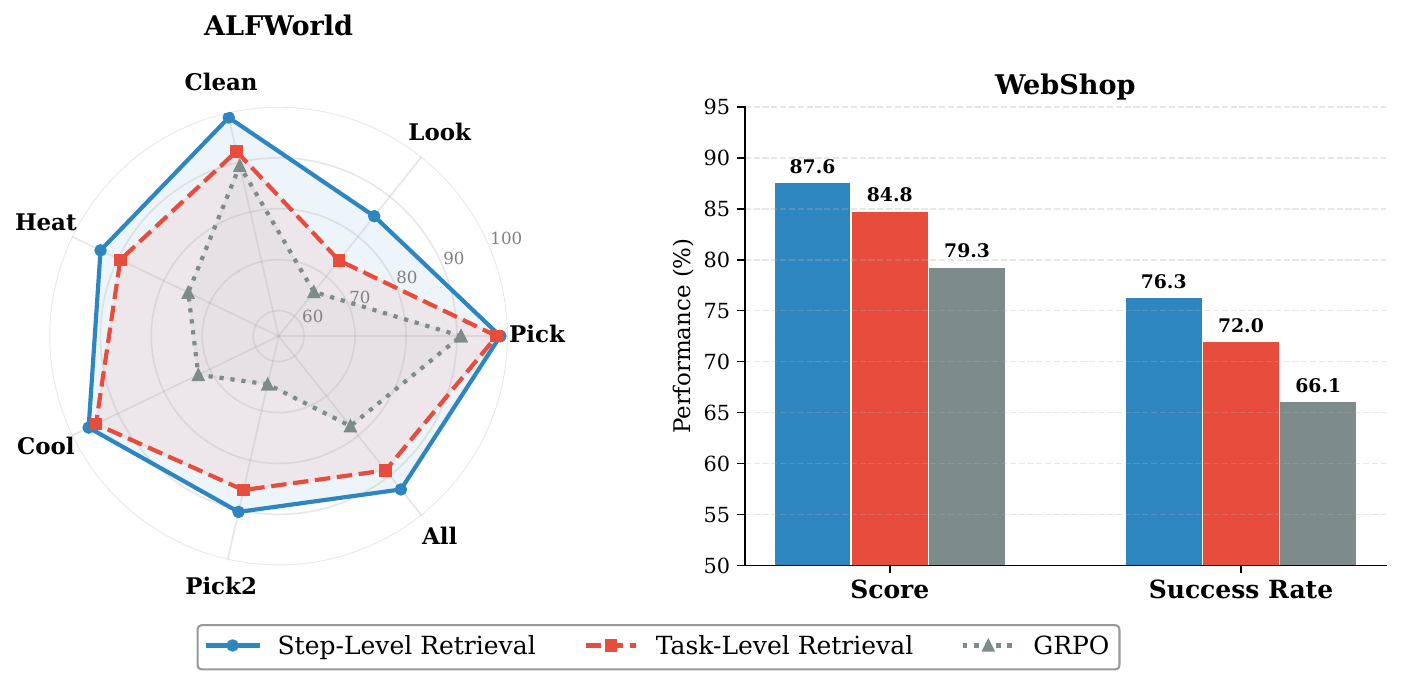}
            \caption{Step-level vs.\ task-level retrieval.}
            \label{fig:ablation-retrieval}
        \end{subfigure}
        \hfill
        \begin{subfigure}[t]{0.49\textwidth}
            \centering
            \includegraphics[width=\textwidth]{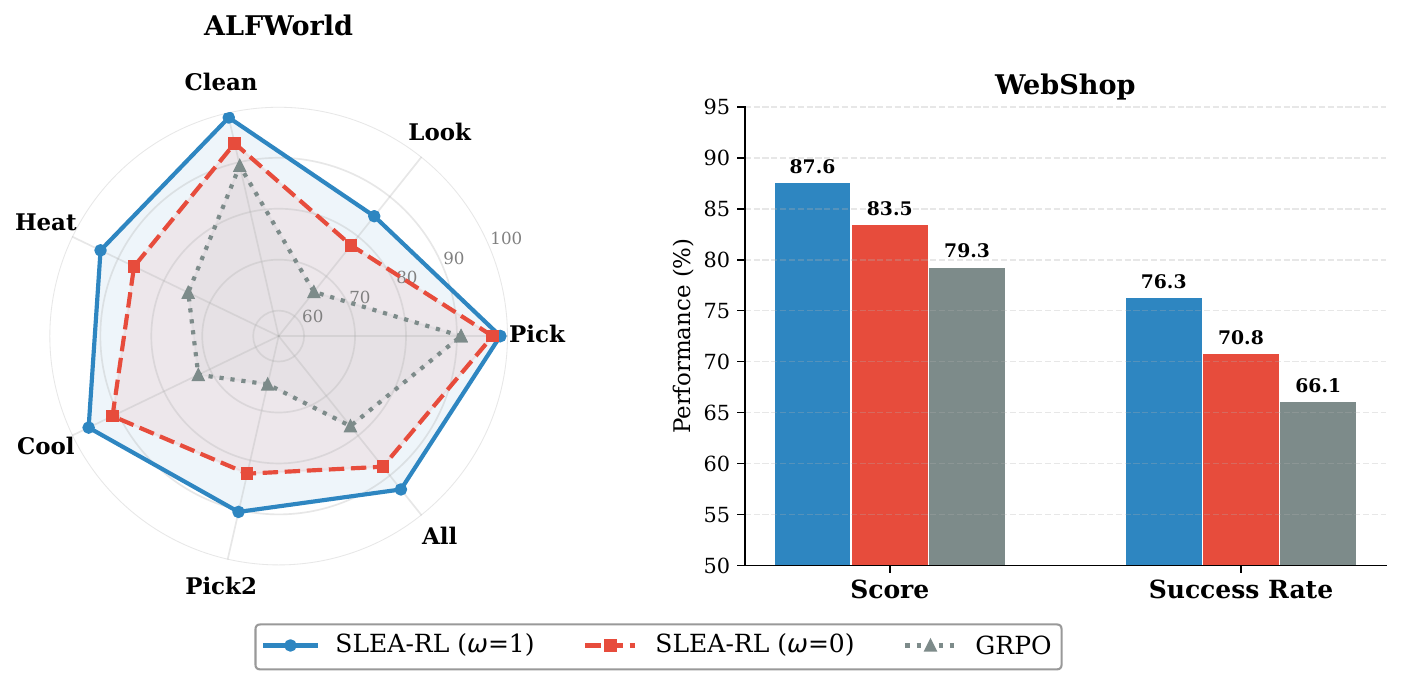}
            \caption{Step-level advantage ($\omega$=0 vs.\ $\omega$=1).}
            \label{fig:ablation-advantage}
        \end{subfigure}
        \caption{Ablation studies on ALFWorld and WebShop (Qwen2.5-7B-Instruct). GRPO is shown as the vanilla RL baseline. (a)~Step-level retrieval consistently outperforms task-level retrieval across all subtasks. (b)~Removing step-level advantage ($\omega$=0) degrades performance below GRPO on several metrics.}
        \label{fig:ablation}
        \vspace{-2mm}
        \end{figure*}

    \subsection{Main Results}

    \paragraph{ALFWorld and WebShop.}
    Table~\ref{tab:alfworld-webshop} reports results on ALFWorld and WebShop. \texttt{SLEA}-RL achieves 93.5\% on ALFWorld and 76.3\% on WebShop, consistently outperforming all baselines. Prompt-based methods plateau well below RL approaches (e.g., ExpeL at 46.3\% on ALFWorld), highlighting the limits of in-context learning for distilling actionable knowledge. Compared to the strongest RL baseline GiGPO (90.8\% ALFWorld, 72.8\% WebShop), \texttt{SLEA}-RL shows gains that are particularly pronounced on harder subtasks requiring multi-step reasoning---Heat improves by 10.1\% and Pick2 by 11.2\%. Among experience-augmented RL methods, \texttt{SLEA}-RL outperforms SkillRL (+3.6\% on both benchmarks), validating that step-level retrieval conditioned on the current observation provides more targeted guidance than task-level skill retrieval. The gains are consistent across model sizes: on Qwen2.5-1.5B-Instruct, \texttt{SLEA}-RL achieves 87.5\% on ALFWorld and 75.4\% on WebShop, with the WebShop improvement (+10.4\% over GiGPO) suggesting that experience augmentation particularly benefits smaller models with less internalized knowledge.

    \paragraph{Search-augmented QA.}
    Table~\ref{tab:search-qa} shows results on search-augmented QA tasks. \texttt{SLEA}-RL achieves the best average score of 60.9\%, outperforming the strongest baseline IGPO (58.7\%). The gains are most apparent on multi-hop tasks requiring iterative information synthesis---\texttt{SLEA}-RL improves over IGPO by 2.3\% on 2Wiki and 9.0\% on MuSiQue (77.2 vs.\ 31.4)---indicating that step-level experiences effectively guide multi-step reasoning. Despite training on only NQ and HotpotQA, \texttt{SLEA}-RL generalizes well to out-of-domain benchmarks (81.8\% on TriviaQA, 55.2\% on PopQA). We note that IGPO achieves strong performance on Bamboogle (74.9\%), where \texttt{SLEA}-RL (33.2\%) lags behind; we attribute this to the dataset's small size and distinctive reasoning patterns that favor IGPO's information-theoretic grouping. Compared to SkillRL, which achieves the best PopQA performance (73.8\%) through task-specific skill retrieval but degrades sharply on multi-hop tasks (20.2\% on MuSiQue), \texttt{SLEA}-RL provides more balanced improvements across both single-hop and multi-hop settings.

    \begin{table*}[t]
    \caption{Performance on ALFWorld and WebShop. We report average success rate (\%) for ALFWorld subtasks and overall, and both average score and success rate (\%) for WebShop. $^*$: results from~\citet{feng2025group}. Best and second best in \textbf{bold} and \underline{underline}. The full table including the prompt-based models is included in Appendix~\ref{sec:appendix_tables}.}
    \label{tab:alfworld-webshop}
    \centering
    \setlength{\tabcolsep}{3.5pt}
    \renewcommand{\arraystretch}{0.92}
    \footnotesize
    \begin{tabular}{@{}ll ccccccc cc@{}}
    \toprule
    & & \multicolumn{7}{c}{\textbf{ALFWorld}} & \multicolumn{2}{c}{\textbf{WebShop}} \\
    \cmidrule(lr){3-9} \cmidrule(lr){10-11}
    \textbf{Type} & \textbf{Method} & Pick & Look & Clean & Heat & Cool & Pick2 & All & Score & Succ. \\
    \midrule
    \multicolumn{11}{@{}l}{\textit{Qwen2.5-7B-Instruct: RL Methods}} \\
    & RLOO$^*$ & 87.6 & 78.2 & 87.3 & 81.3 & 71.9 & 48.9 & 75.5 & 80.3 & 65.7 \\
    & GRPO$^*$ & 90.8 & 66.1 & 89.3 & 74.7 & 72.5 & 64.7 & 77.6 & 79.3 & 66.1 \\
    & GiGPO$^*$ & \underline{97.7} & 82.7 & \underline{98.8} & 83.7 & 89.3 & 79.2 & 90.8 & 84.4 & 72.8 \\
    & MemRL & 62.8 & 38.5 & 22.2 & 12.5 & 8.00 & 0.00 & 21.4 & 29.5 & 9.20 \\
    & Mem0+GRPO & 78.1 & 54.8 & 56.1 & 31.0 & 65.0 & 26.9 & 54.7 & 58.1 & 37.5 \\
    & SimpleMem+GRPO & 89.5 & 36.3 & 60.0 & 50.0 & 64.9 & 26.3 & 62.5 & 67.8 & 46.9 \\
    & SkillRL & 97.9 & 71.4 & 90.0 & \underline{90.0} & \underline{95.5} & \underline{87.5} & \underline{89.9} & \underline{85.2} & \underline{72.7} \\
    \rowcolor{ourscolor} & \texttt{SLEA}-RL & \textbf{98.5} & \textbf{85.1} & \textbf{99.0} & \textbf{93.8} & \textbf{96.4} & \textbf{90.4} & \textbf{93.5} & \textbf{87.6} & \textbf{76.3} \\
    \midrule
    \multicolumn{11}{@{}l}{\textit{Qwen2.5-1.5B-Instruct}} \\
    & GRPO$^*$ & 85.3 & 53.7 & 84.5 & 78.2 & 59.7 & 53.5 & 72.8 & 75.8 & 56.8 \\
    & GiGPO$^*$ & \underline{94.4} & 67.5 & \underline{94.8} & \underline{94.4} & 79.8 & \underline{76.4} & \underline{86.7} & 83.1 & 65.0 \\
    \rowcolor{ourscolor} & \texttt{SLEA}-RL & 92.6 & \textbf{80.0} & 96.2 & 94.7 & \textbf{95.0} & 76.5 & \textbf{87.5} & \textbf{88.7} & \textbf{75.4} \\
    \bottomrule
    \end{tabular}
    \vspace{-3mm}
    \end{table*}

    \begin{table*}[t]
    \caption{Performance on search-augmented QA tasks. \texttt{SLEA}-RL is trained on NQ and HotpotQA. $^\dagger$: in-domain datasets. Best and second best among RL methods in \textbf{bold} and \underline{underline}.}
    \label{tab:search-qa}
    \centering
    \setlength{\tabcolsep}{4pt}
    \renewcommand{\arraystretch}{0.92}
    \footnotesize
    \begin{tabular}{@{}ll ccc c ccc c@{}}
    \toprule
    & & \multicolumn{3}{c}{\textbf{Single-Hop QA}} & & \multicolumn{3}{c}{\textbf{Multi-Hop QA}} & \\
    \cmidrule(lr){3-5} \cmidrule(lr){7-9}
    \textbf{Type} & \textbf{Method} & NQ$^\dagger$ & TriviaQA & PopQA & HotpotQA$^\dagger$ & 2Wiki & MuSiQue & Bamboogle & Avg. \\
    \midrule
    \multicolumn{10}{@{}l}{\textit{Qwen2.5-7B-Instruct}} \\
    & RLOO & 40.7 & 72.5 & 43.1 & 49.6 & 55.0 & 62.2 & 24.8 & 49.7 \\
    & PPO & 38.7 & 75.4 & 48.7 & 48.6 & 59.7 & 63.4 & 26.2 & 51.5 \\
    & GRPO & 40.3 & 77.0 & 49.6 & 48.9 & 57.7 & 65.1 & 25.0 & 51.9 \\
    & Reinforce++ & 34.3 & 67.5 & 44.3 & 45.9 & 54.5 & 61.2 & 23.7 & 47.3 \\
    & GSPO & 41.5 & 77.7 & 45.4 & 46.3 & 60.1 & \underline{67.6} & 25.4 & 52.0 \\
    & GiGPO & 46.4 & 64.7 & 46.1 & 41.6 & 43.6 & 18.9 & \underline{68.9} & 47.2 \\
    & IGPO & \underline{46.7} & \underline{80.1} & 52.5 & \underline{57.2} & \underline{68.2} & 31.4 & \textbf{74.9} & \underline{58.7} \\
    & SkillRL & 45.9 & 63.3 & \textbf{73.8} & 45.9 & 43.2 & 20.2 & 40.3 & 47.1 \\
    \rowcolor{ourscolor} & \texttt{SLEA}-RL & \textbf{48.5} & \textbf{81.8} & \underline{55.2} & \textbf{59.8} & \textbf{70.5} & \textbf{77.2} & 33.2 & \textbf{60.9} \\
    \bottomrule
    \end{tabular}
    \vspace{-2mm}
    \end{table*}


    \subsection{Ablation Studies}


    \paragraph{Task-level vs.\ step-level retrieval.}
    Figure~\ref{fig:ablation-retrieval} compares step-level retrieval (ours) with task-level retrieval, where experiences are retrieved once from the initial task and kept fixed. Step-level retrieval consistently outperforms task-level retrieval across all ALFWorld subtasks, with the largest gains on Look (+11.1\%) and Pick2 (+4.4\%), which require adapting to evolving states. On WebShop, it improves score (87.6 vs.\ 84.8) and success rate (76.3\% vs.\ 72.0\%). Both variants significantly outperform GRPO (77.6\% on ALFWorld, 66.1\% on WebShop), showing that experience augmentation is beneficial, while step-level conditioning provides additional gains. Methodology-wise, the task-level retrieval setting is equivalent to SkillRL.


    \paragraph{Step-level advantage.}
    Figure~\ref{fig:ablation-advantage} evaluates the impact of step-level credit assignment by comparing $w{=}1$ (full method) and $w{=}0$ (episode-level only). Removing step-level advantage reduces ALFWorld performance from 93.5\% to 87.8\%, with larger drops on Look (85.1\% $\to$ 77.8\%) and Heat (93.8\% $\to$ 86.5\%), and lowers WebShop success from 76.3\% to 70.8\%. 
    While $w{=}0$ still outperforms GRPO, step-level advantage provides additional gains (+5.7\% on ALFWorld, +5.5\% on WebShop), indicating that fine-grained credit assignment complements experience retrieval.


    \begin{figure}[t]
    \centering
    \includegraphics[width=0.85\columnwidth]{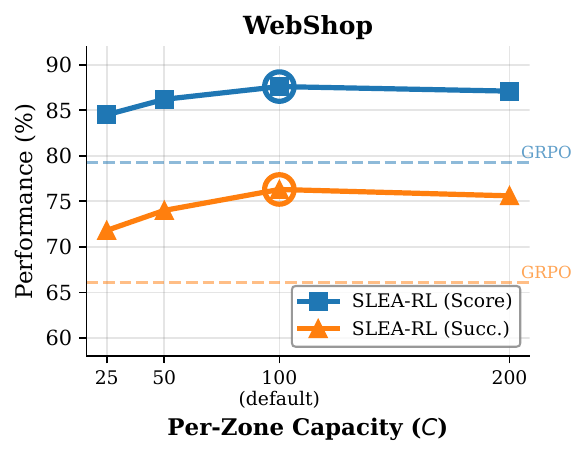}
    \caption{Effect of per-zone capacity ($C$) of the experience library on WebShop. Performance peaks at the default $C$=100 and slightly degrades at $C$=200. All configurations outperform GRPO (dashed lines).}
    \label{fig:ablation-library}
    \vspace{-2mm}
    \end{figure}

    \paragraph{Experience library capacity.}
    Figure~\ref{fig:ablation-library} studies the effect of the per-zone capacity $C$ (strategy and warning) on WebShop. Performance improves from $C{=}25$ (84.5 / 71.8\%) to $C{=}100$ (87.6 / 76.3\%), then slightly declines at $C{=}200$ (87.1 / 75.6\%). This suggests that overly large libraries admit lower-quality experiences that reduce retrieval precision. All settings outperform GRPO (79.3 / 66.1\%), indicating robustness to $C$. We use $C{=}100$ as the default choice.


    \section{Related Work}

    \paragraph{Reinforcement learning for LLM agents.}
    RL has become the dominant paradigm for post-training of LLMs~\citep{schulman2017proximal, bai2022training, rafailov2023direct}, with recent work extending these techniques to multi-turn agentic settings. Group-based methods avoid training a separate critic: RLOO~\citep{ahmadian2024back} uses leave-one-out baselines, GRPO~\citep{guo2025deepseek} computes group-relative advantages, and GSPO~\citep{zheng2025gspo} and Reinforce++~\citep{hu2025reinforcepp} further refine variance reduction. For multi-turn tasks, GiGPO~\citep{feng2025group} provides step-level credit assignment by grouping similar observations, SPEAR~\citep{qin2025learn} employs self-imitation learning, Turn-PPO~\citep{li2025turn} applies per-turn optimization, R$^3$L~\citep{shi2026r} introduces retrospective reward redistribution, and IGPO~\citep{wang2025information} uses information-theoretic grouping.

    \paragraph{Experience-augmented methods for LLM agents.}
    Retrieval-augmented generation~\citep{lewis2020retrieval} enhances LLMs by conditioning on retrieved content. For agentic tasks, prompt-based methods such as ReAct~\citep{yao2022react}, Reflexion~\citep{shinn2023reflexion}, and ExpeL~\citep{zhao2024expel} leverage in-context reasoning or external memory without parameter updates, while Mem0~\citep{chhikara2025mem0}, MemP~\citep{fang2025memp}, and SimpleMem~\citep{liu2026simplemem} maintain persistent memory stores. Training-free experience-based approaches~\citep{cai2025training, flex} achieve agent adaptation through semantically evolving experience libraries without gradient computation. More recently, methods integrating experience with RL training have emerged: \citet{rlep} introduce experience replay during RL, MemRL~\citep{zhang2026memrl} couples memory updates with policy optimization, and SkillRL~\citep{xia2026skillrl} retrieves task-level skills to augment rollouts. However, these approaches retrieve experiences at the task level---once per episode based on the initial description---which becomes stale as observations evolve across steps. A more comprehensive discussion of baseline comparisons is provided in Appendix~\ref{sec:appendix_extended_related}.

\section{Conclusion}
We propose \texttt{SLEA}-RL, a multi-turn reinforcement learning framework that integrates step-level experience retrieval with quality-controlled experience evolution. By combining a self-evolving experience library, cluster-indexed retrieval, and multi-level credit assignment, \texttt{SLEA}-RL enables agents to reuse and refine turn-level knowledge throughout training. Experiments across multiple benchmarks demonstrate consistent improvements over strong baselines. 

\section*{Limitations}

\texttt{SLEA}-RL introduces additional computational overhead due to step-level experience retrieval and prompt augmentation at each decision step, increasing both training and inference cost compared to standard RL. The framework also relies on an external LLM for experience extraction, whose quality can affect the usefulness of the learned experience library; poor extraction may introduce noisy or suboptimal guidance. Moreover, maintaining and updating the experience library adds system complexity, and the benefits may depend on the quality and diversity of collected trajectories. Improving the efficiency and robustness of experience extraction and retrieval remains an important direction for future work.

\bibliography{custom}

@article{achiam2023gpt,
  title={GPT-4 technical report},
  author={Achiam, Josh and Adler, Steven and Agarwal, Sandhini and Ahmad, Lama and Akkaya, Ilge and Aleman, Florencia Leoni and Almeida, Diogo and Altenschmidt, Janko and Altman, Sam and Anadkat, Shyamal and others},
  journal={arXiv preprint arXiv:2303.08774},
  year={2023}
}

@article{team2023gemini,
  title={Gemini: A family of highly capable multimodal models},
  author={Team, Gemini and Anil, Rohan and Borgeaud, Sebastian and Alayrac, Jean-Baptiste and Yu, Jiahui and Soricut, Radu and Schalkwyk, Johan and Dai, Andrew M and Hauth, Anja and Millican, Katie and others},
  journal={arXiv preprint arXiv:2312.11805},
  year={2023}
}

@article{hui2024qwen2,
  title={Qwen2.5-Coder technical report},
  author={Hui, Binyuan and Yang, Jian and Cui, Zeyu and Yang, Jiaxi and Liu, Dayiheng and Zhang, Lei and Liu, Tianyu and Zhang, Jiajun and Yu, Bowen and Lu, Keming and others},
  journal={arXiv preprint arXiv:2409.12186},
  year={2024}
}

@article{liu2024deepseek,
  title={DeepSeek-V3 technical report},
  author={Liu, Aixin and Feng, Bei and Xue, Bing and Wang, Bingxuan and Wu, Bochao and Lu, Chengda and Zhao, Chenggang and Deng, Chengqi and Zhang, Chenyu and Ruan, Chong and others},
  journal={arXiv preprint arXiv:2412.19437},
  year={2024}
}

@article{schulman2017proximal,
  title={Proximal policy optimization algorithms},
  author={Schulman, John and Wolski, Filip and Dhariwal, Prafulla and Radford, Alec and Klimov, Oleg},
  journal={arXiv preprint arXiv:1707.06347},
  year={2017}
}

@article{bai2022training,
  title={Training a helpful and harmless assistant with reinforcement learning from human feedback},
  author={Bai, Yuntao and Jones, Andy and Ndousse, Kamal and Askell, Amanda and Chen, Anna and DasSarma, Nova and Drain, Dawn and Fort, Stanislav and Ganguli, Deep and Henighan, Tom and others},
  journal={arXiv preprint arXiv:2204.05862},
  year={2022}
}

@article{rafailov2023direct,
  title={Direct preference optimization: Your language model is secretly a reward model},
  author={Rafailov, Rafael and Sharma, Archit and Mitchell, Eric and Manning, Christopher D and Ermon, Stefano and Finn, Chelsea},
  journal={Advances in Neural Information Processing Systems},
  volume={36},
  pages={53728--53741},
  year={2023}
}

@inproceedings{ahmadian2024back,
  title={Back to basics: Revisiting REINFORCE-style optimization for learning from human feedback in LLMs},
  author={Ahmadian, Arash and Cremer, Chris and Gall{\'e}, Matthias and Fadaee, Marzieh and Kreutzer, Julia and Pietquin, Olivier and {\"U}st{\"u}n, Ahmet and Hooker, Sara},
  booktitle={Proceedings of the 62nd Annual Meeting of the Association for Computational Linguistics (Volume 1: Long Papers)},
  pages={12248--12267},
  year={2024}
}

@article{guo2025deepseek,
  title={DeepSeek-R1: Incentivizing reasoning capability in {LLMs} via reinforcement learning},
  author={Guo, Daya and Yang, Dejian and Zhang, Haowei and Song, Junxiao and Zhang, Ruoyu and Xu, Runxin and Zhu, Qihao and Ma, Shirong and Wang, Peiyi and Bi, Xiao and others},
  journal={arXiv preprint arXiv:2501.12948},
  year={2025}
}

@article{liu2025understanding,
  title={Understanding r1-zero-like training: A critical perspective},
  author={Liu, Zichen and Chen, Changyu and Li, Wenjun and Qi, Penghui and Pang, Tianyu and Du, Chao and Lee, Wee Sun and Lin, Min},
  journal={arXiv preprint arXiv:2503.20783},
  year={2025}
}

@article{feng2025group,
  title={Group-in-Group Policy Optimization for {LLM} Agent Training},
  author={Feng, Lang and Xue, Zhenghai and Liu, Tingcong and An, Bo},
  journal={arXiv preprint arXiv:2505.10978},
  year={2025}
}

@article{qin2025learn,
  title={Learn the ropes, then trust the wins: self-imitation with progressive exploration for agentic reinforcement learning},
  author={Qin, Yulei and Tan, Xiaoyu and He, Zhengbao and Li, Gang and Lin, Haojia and Li, Zongyi and Xu, Zihan and Shi, Yuchen and Cai, Siqi and Rui, Renting and others},
  journal={arXiv preprint arXiv:2509.22601},
  year={2025}
}

@article{wang2025information,
  title={Information Gain-based Policy Optimization: A Simple and Effective Approach for Multi-Turn LLM Agents},
  author={Wang, Guoqing and Dai, Sunhao and Ye, Guangze and Gan, Zeyu and Yao, Wei and Deng, Yong and Wu, Xiaofeng and Ying, Zhenzhe},
  journal={arXiv preprint arXiv:2510.14967},
  year={2025}
}

@article{shi2026r,
  title={{R\textsuperscript{3}L}: Reflect-then-Retry Reinforcement Learning with Language-Guided Exploration, Pivotal Credit, and Positive Amplification},
  author={Shi, Weijie and Chen, Yanxi and Li, Zexi and Pan, Xuchen and Sun, Yuchang and Xu, Jiajie and Zhou, Xiaofang and Li, Yaliang},
  journal={arXiv preprint arXiv:2601.03715},
  year={2026}
}

@article{li2025turn,
  title={Turn-ppo: Turn-level advantage estimation with ppo for improved multi-turn rl in agentic llms},
  author={Li, Junbo and Zhou, Peng and Meng, Rui and Vadera, Meet P and Li, Lihong and Li, Yang},
  journal={arXiv preprint arXiv:2512.17008},
  year={2025}
}

@article{xia2026skillrl,
  title={SkillRL: Evolving Agents via Recursive Skill-Augmented Reinforcement Learning},
  author={Xia, Peng and Chen, Jianwen and Wang, Hanyang and Liu, Jiaqi and Zeng, Kaide and Wang, Yu and Han, Siwei and Zhou, Yiyang and Zhao, Xujiang and Chen, Haifeng and others},
  journal={arXiv preprint arXiv:2602.08234},
  year={2026}
}

@article{zhang2026memrl,
  title={Memrl: Self-evolving agents via runtime reinforcement learning on episodic memory},
  author={Zhang, Shengtao and Wang, Jiaqian and Zhou, Ruiwen and Liao, Junwei and Feng, Yuchen and Li, Zhuo and Zheng, Yujie and Zhang, Weinan and Wen, Ying and Li, Zhiyu and others},
  journal={arXiv preprint arXiv:2601.03192},
  year={2026}
}

@inproceedings{lewis2020retrieval,
  title={Retrieval-Augmented Generation for Knowledge-Intensive {NLP} Tasks},
  author={Lewis, Patrick and Perez, Ethan and Piktus, Aleksandra and Petroni, Fabio and Karpukhin, Vladimir and Goyal, Naman and K{\"u}ttler, Heinrich and Lewis, Mike and Yih, Wen-tau and Rockt{\"a}schel, Tim and Riedel, Sebastian and Kiela, Douwe},
  booktitle={Advances in Neural Information Processing Systems},
  volume={33},
  pages={9459--9474},
  year={2020}
}

@misc{rlep,
      title={RLEP: Reinforcement Learning with Experience Replay for LLM Reasoning}, 
      author={Hongzhi Zhang and Jia Fu and Jingyuan Zhang and Kai Fu and Qi Wang and Fuzheng Zhang and Guorui Zhou},
      year={2025},
      eprint={2507.07451},
      archivePrefix={arXiv},
      primaryClass={cs.CL},
      url={https://arxiv.org/abs/2507.07451}, 
}

@article{flex,
  title={Flex: Continuous agent evolution via forward learning from experience},
  author={Cai, Zhicheng and Guo, Xinyuan and Pei, Yu and Feng, Jiangtao and Su, Jinsong and Chen, Jiangjie and Zhang, Ya-Qin and Ma, Wei-Ying and Wang, Mingxuan and Zhou, Hao},
  journal={arXiv preprint arXiv:2511.06449},
  year={2025}
}

@article{cai2025training,
  title={Training-free group relative policy optimization},
  author={Cai, Yuzheng and Cai, Siqi and Shi, Yuchen and Xu, Zihan and Chen, Lichao and Qin, Yulei and Tan, Xiaoyu and Li, Gang and Li, Zongyi and Lin, Haojia and others},
  journal={arXiv preprint arXiv:2510.08191},
  year={2025}
}

@inproceedings{ALFWorld20,
               title ={{ALFWorld: Aligning Text and Embodied
               Environments for Interactive Learning}},
               author={Mohit Shridhar and Xingdi Yuan and
               Marc-Alexandre C\^ot\'e and Yonatan Bisk and
               Adam Trischler and Matthew Hausknecht},
               booktitle = {Proceedings of the International
               Conference on Learning Representations (ICLR)},
               year = {2021},
               url = {https://arxiv.org/abs/2010.03768}}

@article{yao2022webshop,
  title={Webshop: Towards scalable real-world web interaction with grounded language agents},
  author={Yao, Shunyu and Chen, Howard and Yang, John and Narasimhan, Karthik},
  journal={Advances in Neural Information Processing Systems},
  volume={35},
  pages={20744--20757},
  year={2022}
}

@inproceedings{yao2022react,
  title={ReAct: Synergizing Reasoning and Acting in Language Models},
  author={Yao, Shunyu and Zhao, Jeffrey and Yu, Dian and Du, Nan and Shafran, Izhak and Narasimhan, Karthik and Cao, Yuan},
  booktitle={The Eleventh International Conference on Learning Representations},
  year={2022b}
}

@inproceedings{shinn2023reflexion,
  title={Reflexion: Language Agents with Verbal Reinforcement Learning},
  author={Shinn, Noah and Cassano, Federico and Gopinath, Ashwin and Narasimhan, Karthik and Yao, Shunyu},
  booktitle={Advances in Neural Information Processing Systems},
  volume={36},
  pages={8634--8652},
  year={2023}
}

@article{chhikara2025mem0,
  title={Mem0: Building production-ready ai agents with scalable long-term memory},
  author={Chhikara, Prateek and Khant, Dev and Aryan, Saket and Singh, Taranjeet and Yadav, Deshraj},
  journal={arXiv preprint arXiv:2504.19413},
  year={2025}
}

@inproceedings{zhao2024expel,
  title={Expel: {LLM} Agents are Experiential Learners},
  author={Zhao, Andrew and Huang, Daniel and Xu, Quentin and Lin, Matthieu and Liu, Yong-Jin and Huang, Gao},
  booktitle={Proceedings of the AAAI Conference on Artificial Intelligence},
  volume={38},
  pages={19632--19642},
  year={2024}
}

@article{fang2025memp,
  title={Memp: Exploring agent procedural memory},
  author={Fang, Runnan and Liang, Yuan and Wang, Xiaobin and Wu, Jialong and Qiao, Shuofei and Xie, Pengjun and Huang, Fei and Chen, Huajun and Zhang, Ningyu},
  journal={arXiv preprint arXiv:2508.06433},
  year={2025}
}

@article{liu2026simplemem,
  title={SimpleMem: Efficient lifelong memory for {LLM} agents},
  author={Liu, Jiaqi and Su, Yaofeng and Xia, Peng and Han, Siwei and Zheng, Zeyu and Xie, Cihang and Ding, Mingyu and Yao, Huaxiu},
  journal={arXiv preprint arXiv:2601.02553},
  year={2026}
}

@article{wu2025evolver,
  title={EvolveR: Self-evolving {LLM} agents through an experience-driven lifecycle},
  author={Wu, Rong and Wang, Xiaoman and Mei, Jianbiao and Cai, Pinlong and Fu, Daocheng and Yang, Cheng and Wen, Licheng and Yang, Xuemeng and Shen, Yufan and Wang, Yuxin and Shi, Botian},
  journal={arXiv preprint arXiv:2510.16079},
  year={2025}
}

@article{zheng2025gspo,
  title={Group sequence policy optimization},
  author={Zheng, Chujie and Liu, Shixuan and Li, Mingze and Chen, Xiong-Hui and Yu, Bowen and Gao, Chang and Dang, Kai and Liu, Yuqiong and Men, Rui and Yang, An and others},
  journal={arXiv preprint arXiv:2507.18071},
  year={2025}
}

@article{hu2025reinforcepp,
  title={{REINFORCE++}: Stabilizing critic-free policy optimization with global advantage normalization},
  author={Hu, Jian and Liu, Jason Klein and Xu, Haotian and Shen, Wei},
  journal={arXiv preprint arXiv:2501.03262},
  year={2025}
}

@article{jin2025searchr1,
  title={Search-R1: Training {LLMs} to reason and leverage search engines with reinforcement learning},
  author={Jin, Bowen and Zeng, Hansi and Yue, Zhenrui and Yoon, Jinsung and Arik, Sercan and Wang, Dong and Zamani, Hamed and Han, Jiawei},
  journal={arXiv preprint arXiv:2503.09516},
  year={2025}
}

@article{sun2025zerosearch,
  title={ZeroSearch: Incentivize the search capability of {LLMs} without searching},
  author={Sun, Hao and Qiao, Zile and Guo, Jiayan and Fan, Xuanbo and Hou, Yingyan and Jiang, Yong and Xie, Pengjun and Huang, Fei and Zhang, Yan},
  journal={arXiv preprint arXiv:2505.04588},
  year={2025}
}

@inproceedings{zheng2025stepsearch,
  title={{StepSearch}: Igniting {LLMs} search ability via step-wise proximal policy optimization},
  author={Zheng, Xuhui and An, Kang and Wang, Ziliang and Wang, Yuhang and Wu, Yichao},
  booktitle={Proceedings of the 2025 Conference on Empirical Methods in Natural Language Processing},
  pages={21816--21841},
  year={2025}
}

@article{kwiatkowski2019natural,
  title={Natural Questions: A benchmark for question answering research},
  author={Kwiatkowski, Tom and Palomaki, Jennimaria and Redfield, Olivia and Collins, Michael and Parikh, Ankur and Alberti, Chris and Epstein, Danielle and Polosukhin, Illia and Devlin, Jacob and Lee, Kenton and others},
  journal={Transactions of the Association for Computational Linguistics},
  volume={7},
  pages={452--466},
  year={2019}
}

@inproceedings{joshi2017triviaqa,
  title={TriviaQA: A large scale distantly supervised challenge dataset for reading comprehension},
  author={Joshi, Mandar and Choi, Eunsol and Weld, Daniel S. and Zettlemoyer, Luke},
  booktitle={Proceedings of the 55th Annual Meeting of the Association for Computational Linguistics (Volume 1: Long Papers)},
  pages={1601--1611},
  year={2017}
}

@inproceedings{mallen2023trust,
  title={When not to trust language models: Investigating effectiveness of parametric and non-parametric memories},
  author={Mallen, Alex and Asai, Akari and Zhong, Victor and Das, Rajarshi and Khashabi, Daniel and Hajishirzi, Hannaneh},
  booktitle={Proceedings of the 61st Annual Meeting of the Association for Computational Linguistics (Volume 1: Long Papers)},
  pages={9802--9822},
  year={2023}
}

@inproceedings{yang2018hotpotqa,
  title={{HotpotQA}: A dataset for diverse, explainable multi-hop question answering},
  author={Yang, Zhilin and Qi, Peng and Zhang, Saizheng and Bengio, Yoshua and Cohen, William W. and Salakhutdinov, Ruslan and Manning, Christopher D.},
  booktitle={Proceedings of the 2018 Conference on Empirical Methods in Natural Language Processing},
  pages={2369--2380},
  year={2018}
}

@inproceedings{ho2020constructing,
  title={Constructing a multi-hop {QA} dataset for comprehensive evaluation of reasoning steps},
  author={Ho, Xanh and Nguyen, Anh-Khoa Duong and Sugawara, Saku and Aizawa, Akiko},
  booktitle={Proceedings of the 28th International Conference on Computational Linguistics},
  pages={6609--6625},
  year={2020}
}

@article{trivedi2022musique,
  title={{MuSiQue}: Multihop questions via single-hop question composition},
  author={Trivedi, Harsh and Balasubramanian, Niranjan and Khot, Tushar and Sabharwal, Ashish},
  journal={Transactions of the Association for Computational Linguistics},
  volume={10},
  pages={539--554},
  year={2022}
}

@inproceedings{press2023measuring,
  title={Measuring and narrowing the compositionality gap in language models},
  author={Press, Ofir and Zhang, Muru and Min, Sewon and Schmidt, Ludwig and Smith, Noah A. and Lewis, Mike},
  booktitle={Findings of the Association for Computational Linguistics: EMNLP 2023},
  pages={5687--5711},
  year={2023}
}

\clearpage
\appendix

\section{Format-Preserving Prompt Augmentation}
\label{sec:format-preserving}

A critical implementation detail concerns how retrieved experiences are incorporated into the prompt. Naively prepending experience text directly into the observation content disrupts the format the model expects, causing progressive degradation of valid action generation as the library grows. The policy $\pi_\theta$ is well-calibrated for observations in the expected clean format (i.e., the observation distribution $\mathcal{O}_{\text{clean}}$ seen during pretraining and supervised fine-tuning); when the observation is corrupted with arbitrary preamble text, the distribution shift increases with the length of the injected text, producing increasingly malformed actions.

\texttt{SLEA}-RL addresses this through format-preserving augmentation. When experiences $\varepsilon_t$ are available, the prompt is constructed as:
\begin{equation}
\text{augment}(o_t, \varepsilon_t) = [\text{sys}(\varepsilon_t),\ \text{obs}(o_t)]
\end{equation}
where $\text{sys}(\varepsilon_t)$ places the retrieved strategies and warnings in the system role and $\text{obs}(o_t)$ preserves the observation in the user role exactly as the model expects. This separation ensures that experience injection acts as a small perturbation to the policy rather than a format-breaking corruption. The experience text is constrained to a token budget $B_{\max}$ to avoid consuming excessive prompt context.

\section{Analysis: Step-Level vs.\ Task-Level Retrieval}

The experience library reduces uncertainty about optimal actions at each decision point. Letting $H(a^*_t \mid o_t)$ denote the entropy of the optimal action distribution at step $t$ given observation $o_t$, access to relevant experiences $\varepsilon$ provides information gain:
\begin{equation}
I(a^*_t; \varepsilon \mid o_t) = H(a^*_t \mid o_t) - H(a^*_t \mid o_t, \varepsilon) > 0
\end{equation}
The key distinction lies in what experiences are retrieved. Task-level retrieval conditions on the initial task description $d$, retrieving $\varepsilon_{\text{task}} = \text{retrieve}(E, d)$ once for the entire episode. Step-level retrieval conditions on the current observation, retrieving $\varepsilon_t = \text{retrieve}(E, o_t)$ at each step. Since $o_t$ provides more specific context about the agent's current situation than the initial task description, step-level retrieval is expected to yield more relevant experiences:
\begin{equation}
I(a^*_t; \varepsilon_t \mid o_t) \geq I(a^*_t; \varepsilon_{\text{task}} \mid o_t)
\end{equation}
This advantage becomes increasingly pronounced as the episode progresses and $o_t$ diverges from the initial context. By step 20 of a household navigation task, the agent's current room and available objects bear little resemblance to the original goal specification, rendering task-level experiences increasingly irrelevant. We validate this hypothesis empirically through ablation studies comparing step-level, task-level, and hybrid retrieval modes.

\section{More About Benchmarks}
\label{sec:appendix_benchmarks}
\textit{ALFWorld}~\citep{ALFWorld20} is an embodied decision-making benchmark where agents must complete multi-step household tasks through textual interactions with a simulated environment. Tasks require long-horizon planning, state tracking, and grounding actions in partially observable settings.

\textit{WebShop}~\citep{yao2022webshop} is a web-based interactive environment that evaluates agents in realistic online shopping scenarios. The agent navigates a simulated HTML interface to search for, filter, and purchase products based on natural language instructions. The environment contains over 1.1M products and 12K instructions, yielding a large and diverse action space.

We further evaluate on \textit{search-augmented QA tasks}, which require agents to iteratively retrieve and reason over external information. These include \textit{single-hop} datasets (NQ~\citep{kwiatkowski2019natural}, TriviaQA~\citep{joshi2017triviaqa}, PopQA~\citep{mallen2023trust}), where answers can typically be found within a single document, and \textit{multi-hop} datasets (HotpotQA~\citep{yang2018hotpotqa}, 2Wiki~\citep{ho2020constructing}, MuSiQue~\citep{trivedi2022musique}, Bamboogle~\citep{press2023measuring}), which require aggregating evidence across multiple sources. These benchmarks test both retrieval quality and multi-step reasoning.


\section{Hyperparameter Configuration}
\label{sec:hyperparams}

Table~\ref{tab:hyperparams} lists the full hyperparameter configuration for \texttt{SLEA}-RL.

\begin{table}[h]
\begin{center}
\begin{tabular}{ll}
\toprule
\textbf{Hyperparameter} & \textbf{Value} \\
\midrule
\multicolumn{2}{l}{\textit{Step-Level Retrieval}} \\
Retrieval mode & clustered \\
Top-$k$ golden strategies & 2 \\
Top-$k$ warnings & 1 \\
Max experience tokens & 200 \\
Similarity threshold $\delta$ & 0.85 \\
\midrule
\multicolumn{2}{l}{\textit{Warmup and Quality Gates}} \\
Warmup epochs $W$ & 5 \\
Min library size $C_{\min}$ & 10 \\
\midrule
\multicolumn{2}{l}{\textit{Self-Evolving Experience}} \\
Golden capacity per level $C$ & 100 \\
Warning capacity per level & 50 \\
Max strategies per step $K_{\text{strat}}$ & 10 \\
Max warnings per step $K_{\text{warn}}$ & 5 \\
Top-$k$ trajectories $K_{\text{traj}}$ & 5 \\
Novelty threshold & 0.85 \\
\midrule
\multicolumn{2}{l}{\textit{Policy Optimization (GiGPO)}} \\
Step advantage weight $w$ & 1.0 \\
Advantage mode & mean\_norm \\
\bottomrule
\end{tabular}
\end{center}
\caption{\texttt{SLEA}-RL hyperparameter configuration.}
\label{tab:hyperparams}
\end{table}

\section{Training curves.}
\label{sec:appendix_add_res}

\begin{figure*}[!htbp]
\centering
\includegraphics[width=\textwidth]{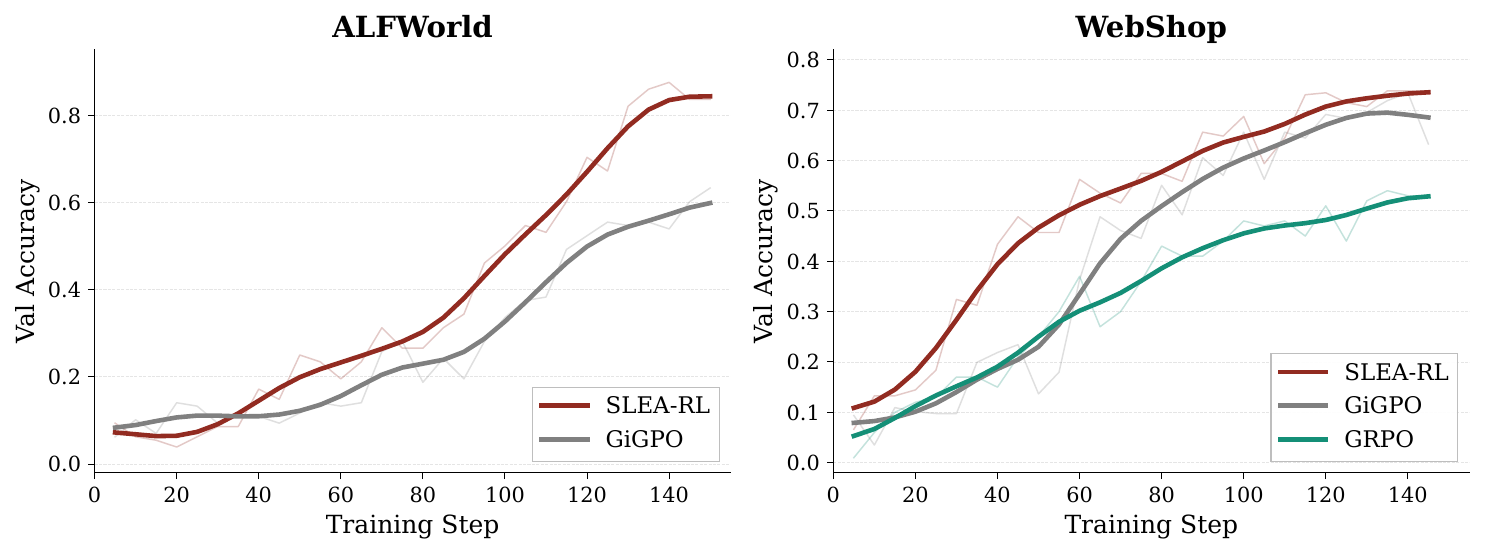}
\caption{Validation success rates on ALFWorld (left) and WebShop (right) with Qwen2.5-1.5B-Instruct. Faint traces show raw values; bold lines are smoothed. \texttt{SLEA}-RL achieves faster convergence and higher asymptotic performance than GiGPO and GRPO.}
\label{fig:training-curves}
\vspace{-2mm}
\end{figure*}

Figure~\ref{fig:training-curves} shows the validation success rates on ALFWorld and WebShop with Qwen2.5-1.5B-Instruct. \texttt{SLEA}-RL achieves both faster convergence and higher final performance compared to GiGPO and GRPO. On ALFWorld, \texttt{SLEA}-RL begins to separate from GiGPO around step 50 and reaches over 85\% success by step 140, while GiGPO plateaus near 63\%. On WebShop, the separation is similarly pronounced: \texttt{SLEA}-RL converges to approximately 74\% success, compared to 72\% for GiGPO and 53\% for GRPO.

\begin{figure*}[!htbp]
\centering
\begin{subfigure}[t]{0.49\textwidth}
    \centering
    \includegraphics[width=\textwidth]{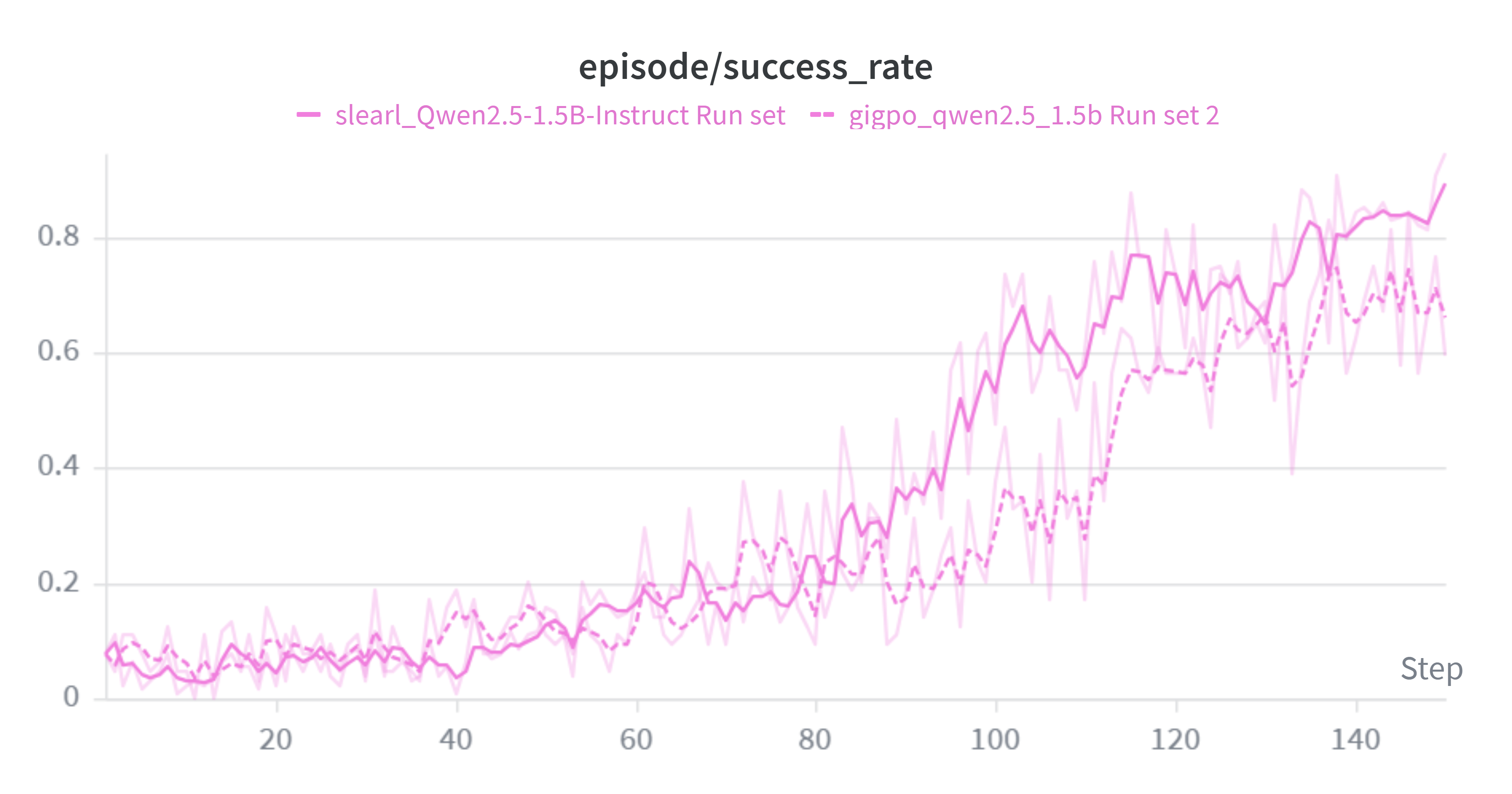}
    \caption{Training success rate (ALFWorld).}
    \label{fig:episode-success-alf}
\end{subfigure}
\hfill
\begin{subfigure}[t]{0.49\textwidth}
    \centering
    \includegraphics[width=\textwidth]{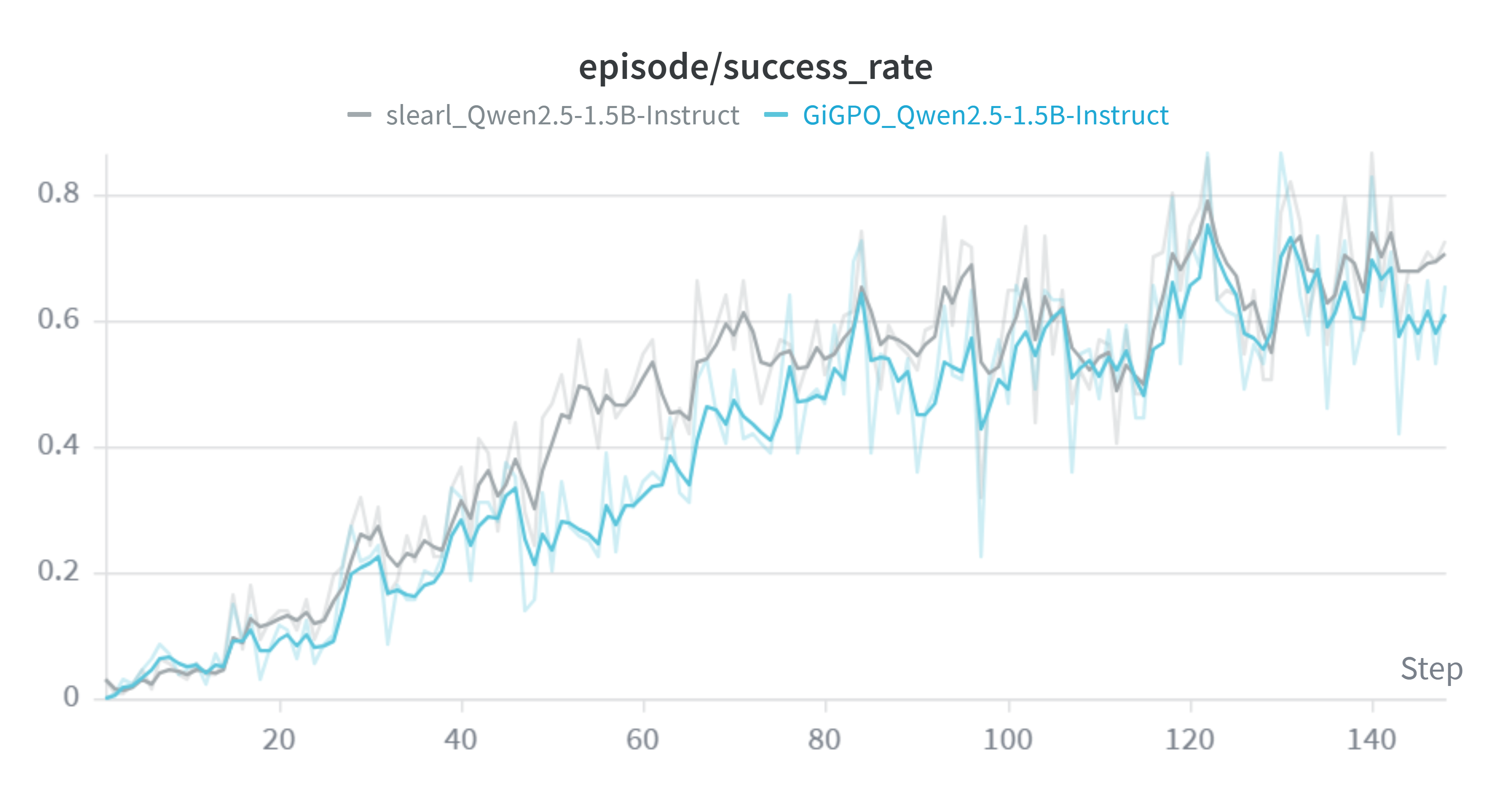}
    \caption{Training success rate (WebShop).}
    \label{fig:episode-success-web}
\end{subfigure}

\vspace{2mm}

\begin{subfigure}[t]{0.49\textwidth}
    \centering
    \includegraphics[width=\textwidth]{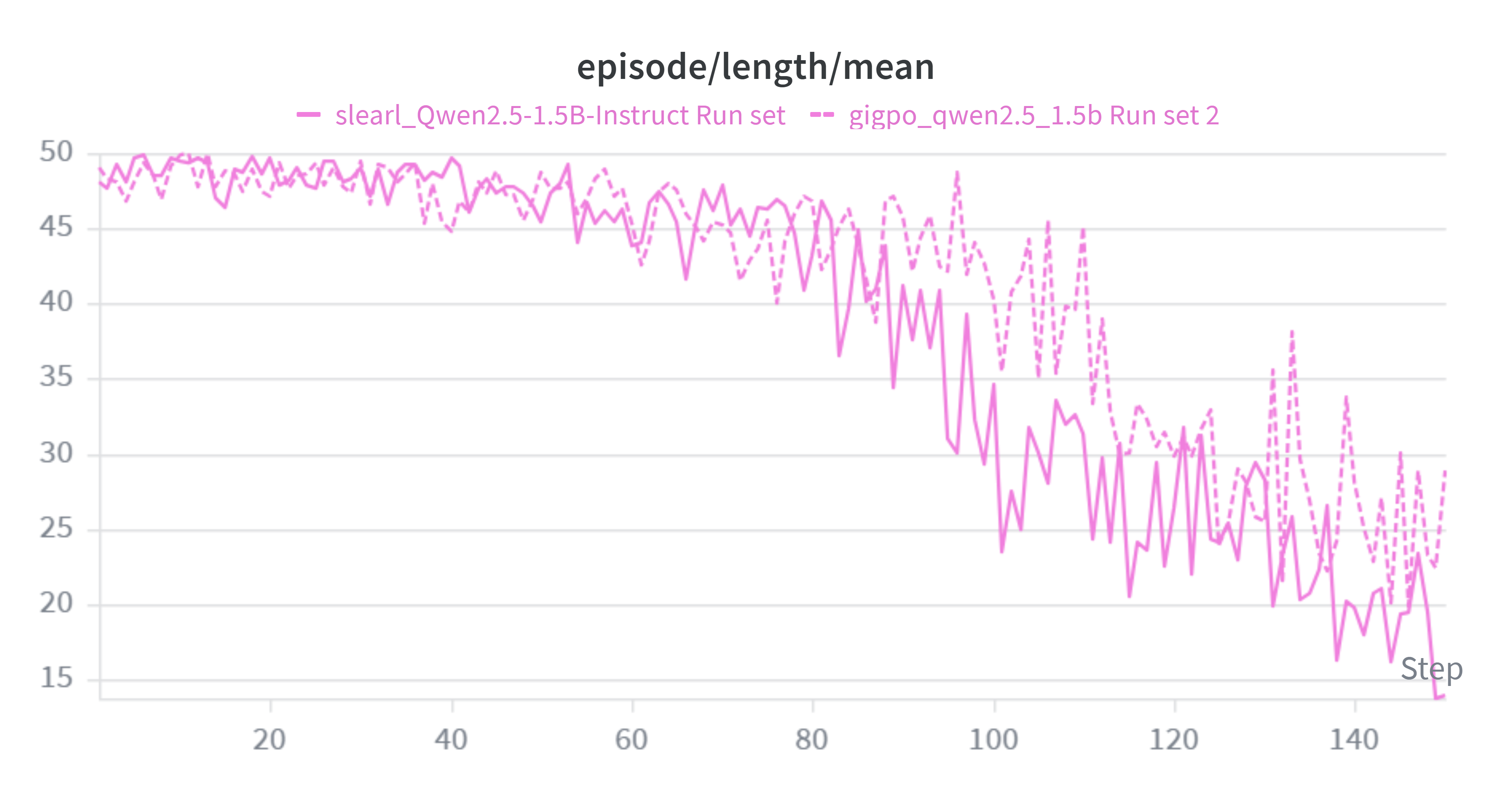}
    \caption{Mean episode length (ALFWorld).}
    \label{fig:episode-length-alf}
\end{subfigure}
\hfill
\begin{subfigure}[t]{0.49\textwidth}
    \centering
    \includegraphics[width=\textwidth]{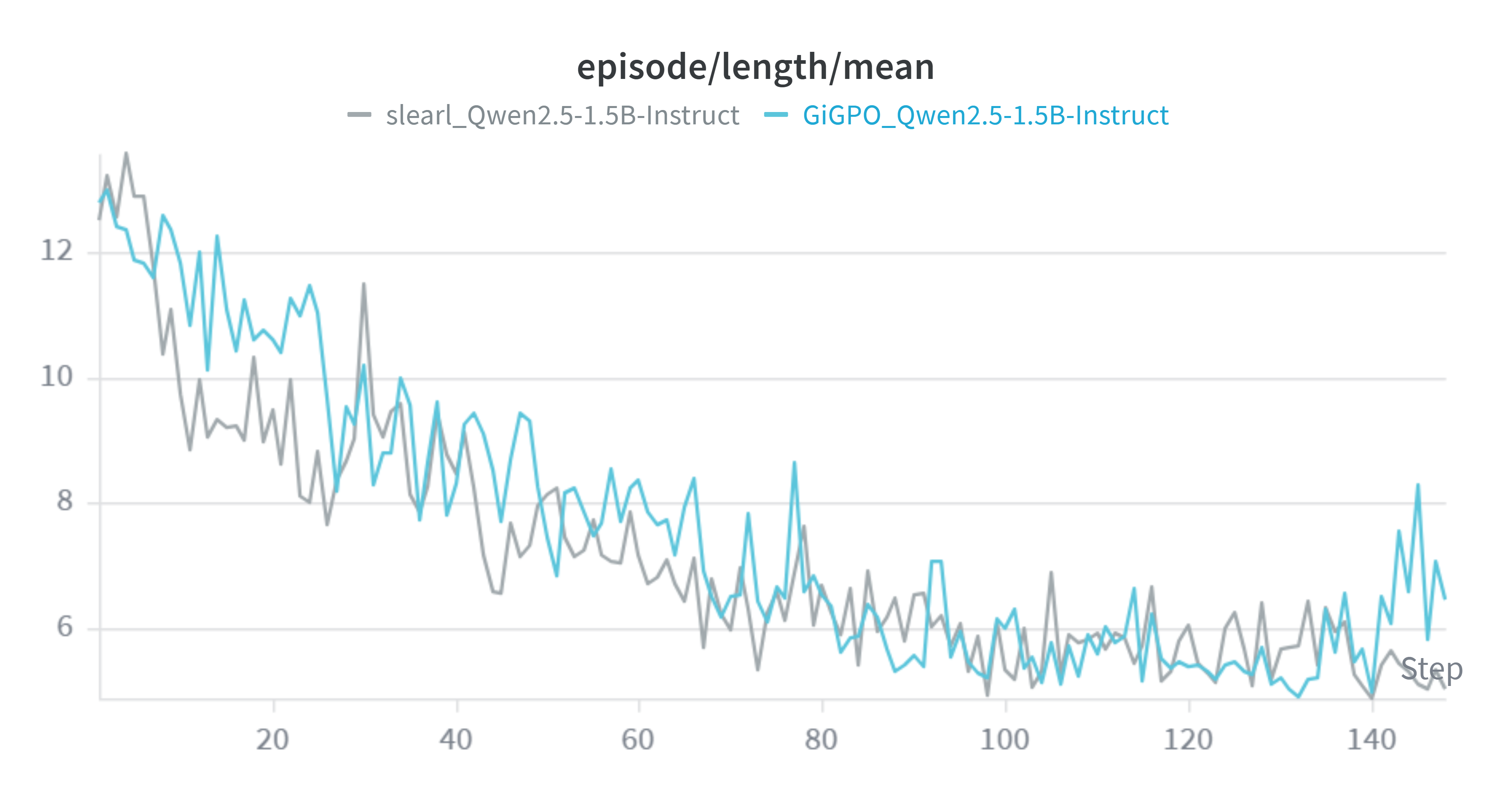}
    \caption{Mean episode length (WebShop).}
    \label{fig:episode-length-web}
\end{subfigure}
\caption{
  Training dynamics on ALFWorld (left) and WebShop (right) with Qwen2.5-1.5B-Instruct. \textbf{Top row}: training episode success rate. \texttt{SLEA}-RL achieves higher success than GiGPO throughout training, with the gap widening as the experience library matures. \textbf{Bottom row}: mean episode length. \texttt{SLEA}-RL reduces episode length faster, indicating more efficient task completion via experience-guided action selection.}
\label{fig:episode-dynamics}
\vspace{-2mm}
\end{figure*}

\paragraph{Training episode success rate.}
Figures~\ref{fig:episode-success-alf} and~\ref{fig:episode-success-web} show the training episode success rate on ALFWorld and WebShop. On ALFWorld, \texttt{SLEA}-RL exhibits a steeper rise starting around step 60, reaching over 80\% by step 130, while GiGPO trails by 10--15\%. On WebShop, \texttt{SLEA}-RL pulls ahead around step 40 and reaches 70--75\% success by step 140, compared to GiGPO's 60\%. The widening gap in the mid-to-late training regime confirms that experience retrieval provides increasingly useful guidance as the library accumulates higher-quality entries.

\paragraph{Episode length dynamics.}
Figures~\ref{fig:episode-length-alf} and~\ref{fig:episode-length-web} show the mean episode length. A decreasing length indicates more efficient task completion. On ALFWorld, both methods begin near 48 steps (the maximum); \texttt{SLEA}-RL drops below 20 by step 140 while GiGPO remains around 25--30, suggesting that retrieved strategies help the agent avoid unnecessary exploration. On WebShop, both converge to 5--6 steps, but \texttt{SLEA}-RL reaches this efficiency earlier, particularly in the mid-training regime (steps 30--80).

\FloatBarrier
\section{System Prompts}
\label{sec:appendix_prompts}

Figure~\ref{fig:system-prompts} presents the three system prompts used in the \texttt{SLEA}-RL pipeline. \textbf{Prompt A} is prepended to the agent's observation at each step when experience retrieval is active, providing retrieved strategies and warnings via format-preserving augmentation (Section~\ref{sec:format-preserving}). \textbf{Prompt B} is used during the experience evolution phase to extract reusable strategies from successful trajectories, organized into three abstraction levels (principles, methods, and concrete examples). \textbf{Prompt C} performs analogous failure diagnosis on unsuccessful trajectories, identifying root causes and formulating actionable warnings. Together, these prompts enable semantic library evolution without gradient updates.

\definecolor{prompt-bg}{RGB}{248,244,255}
\definecolor{prompt-border}{RGB}{140,100,180}
\definecolor{system-bg}{RGB}{255,240,240}
\definecolor{system-border}{RGB}{200,80,80}

\begin{figure*}[t]
\centering
\small

\begin{tcolorbox}[
  colback=white, colframe=black!50,
  fonttitle=\bfseries\large,
  boxrule=0.8pt, arc=3pt, left=6pt, right=6pt, top=4pt, bottom=4pt,
  title={\texttt{SLEA}-RL System Prompts}
]

\begin{tcolorbox}[
  colback=prompt-bg, colframe=prompt-border!50,
  fonttitle=\bfseries\small\color{prompt-border},
  boxrule=0.6pt, arc=3pt, left=4pt, right=4pt, top=3pt, bottom=3pt,
  title={Prompt A: Step-Level Experience Augmentation (prepended to observation at each step)}
]
\ttfamily\footnotesize
Below are relevant strategies and warnings retrieved from past experience. Use them to inform your decisions, but always prioritize the current observation.\\[4pt]
\textnormal{\textbf{[Helpful Strategies]}}\\
\{retrieved\_golden\_strategies\}\\[4pt]
\textnormal{\textbf{[Common Pitfalls]}}\\
\{retrieved\_warnings\}\\[4pt]
\textnormal{\rule{\linewidth}{0.3pt}}\\
\textnormal{\textit{Now proceed with the task. Your current observation follows:}}\\
\{current\_observation\}
\end{tcolorbox}

\vspace{6pt}

\begin{tcolorbox}[
  colback=system-bg, colframe=system-border!50,
  fonttitle=\bfseries\small\color{system-border},
  boxrule=0.6pt, arc=3pt, left=4pt, right=4pt, top=3pt, bottom=3pt,
  title={Prompt B: Strategy Extraction (applied to successful trajectories)}
]
\ttfamily\footnotesize
You are analyzing a \textbf{successful} agent trajectory. The agent completed the task and received a reward of \{reward\}.\\[4pt]
\textnormal{\textbf{Trajectory:}}\\
\{trajectory\_text\}\\[4pt]
\textnormal{\textbf{Instructions:}}\\
1. Identify the key decisions that led to success.\\
2. Abstract them into reusable strategies at three levels:\\
\quad -- \textbf{Principle}: High-level guideline (e.g., ``Always check the most likely location first'')\\
\quad -- \textbf{Method}: Reasoning pattern (e.g., ``For heating tasks: find item $\to$ pick up $\to$ heat $\to$ place'')\\
\quad -- \textbf{Example}: Concrete fact (e.g., ``Eggs are commonly found in the fridge'')\\
3. Format each strategy as a concise, self-contained sentence.\\
4. Only extract strategies that would \textbf{generalize} to similar situations.
\end{tcolorbox}

\vspace{6pt}

\begin{tcolorbox}[
  colback=system-bg, colframe=system-border!50,
  fonttitle=\bfseries\small\color{system-border},
  boxrule=0.6pt, arc=3pt, left=4pt, right=4pt, top=3pt, bottom=3pt,
  title={Prompt C: Failure Diagnosis (applied to failed trajectories)}
]
\ttfamily\footnotesize
You are analyzing a \textbf{failed} agent trajectory. The agent did not complete the task and received a reward of \{reward\}.\\[4pt]
\textnormal{\textbf{Trajectory:}}\\
\{trajectory\_text\}\\[4pt]
\textnormal{\textbf{Instructions:}}\\
1. Identify \textbf{where} in the trajectory the agent went wrong.\\
2. Determine the \textbf{root cause} of the failure (e.g., inefficient search, wrong assumption, repeated action).\\
3. Formulate actionable warnings at three levels:\\
\quad -- \textbf{Mistake}: Specific action to avoid (e.g., ``Do not search cabinets exhaustively without checking fridge first'')\\
\quad -- \textbf{Pattern}: Recurring failure mode (e.g., ``Repeating the same failed action leads to step limit exhaustion'')\\
\quad -- \textbf{Diagnostic}: Root cause insight (e.g., ``Task failed because the agent lacked a mental model of common item locations'')\\
4. Each warning should be concise and self-contained.
\end{tcolorbox}

\end{tcolorbox}

\caption{System prompts used in the \texttt{SLEA}-RL pipeline. \textbf{Prompt A} augments each step's observation with retrieved experiences. \textbf{Prompt B} extracts reusable strategies from successful trajectories into the strategy zone $E^+$. \textbf{Prompt C} diagnoses failures and populates the warning zone $E^-$. These prompts enable semantic library evolution without gradient updates.}
\label{fig:system-prompts}
\end{figure*}

\FloatBarrier
\section{Extended Related Work}
\label{sec:appendix_extended_related}

This section provides a more comprehensive discussion of the relationship between \texttt{SLEA}-RL and existing methods.

\paragraph{Comparison with SkillRL.}
SkillRL~\citep{xia2026skillrl} is the most closely related experience-augmented RL method. It maintains a skill library of reusable strategies and retrieves relevant skills to augment the agent's prompt during RL training. However, SkillRL operates at the \emph{task level}: skills are retrieved once based on the initial task description and held constant throughout the episode. This design is adequate for tasks where the context remains stable, but becomes a limitation in multi-turn environments where the agent's observation evolves at each step. For instance, in ALFWorld, the agent may begin in a kitchen but navigate to a bedroom mid-episode; task-level skills about ``kitchen tasks'' become irrelevant. \texttt{SLEA}-RL addresses this through step-level retrieval conditioned on the current observation $o_t$, ensuring that retrieved experiences remain relevant as the episode progresses. Additionally, SkillRL does not incorporate step-level credit assignment, relying solely on episode-level advantages. \texttt{SLEA}-RL combines step-level retrieval with GiGPO-style step-level advantage estimation, providing fine-grained credit that identifies which specific actions benefit from which experiences.

\paragraph{Comparison with MemRL.}
MemRL~\citep{zhang2026memrl} integrates memory mechanisms directly into the RL optimization loop. Unlike \texttt{SLEA}-RL, which maintains an external experience library that evolves through semantic analysis, MemRL updates its memory bank through gradient-based optimization. While this approach allows end-to-end training, it conflates memory quality with policy quality---if the policy degrades, memory updates may also degrade. \texttt{SLEA}-RL decouples these concerns: the experience library evolves through quality-controlled semantic extraction (score-based admission, novelty checking), independent of gradient updates, providing more stable knowledge accumulation.

\paragraph{Comparison with prompt-based memory methods.}
Methods such as Mem0~\citep{chhikara2025mem0}, ExpeL~\citep{zhao2024expel}, MemP~\citep{fang2025memp}, and SimpleMem~\citep{liu2026simplemem} maintain external memory stores that guide agent behavior through in-context retrieval. These approaches operate without parameter updates, relying entirely on the base model's ability to leverage retrieved context. While effective for rapid adaptation, they cannot fundamentally improve the policy's decision-making capabilities. Hybrid approaches (Mem0+GRPO, SimpleMem+GRPO) combine memory with RL training but still retrieve at the task level. \texttt{SLEA}-RL differs by (i) performing step-level retrieval conditioned on the evolving observation, (ii) using quality-controlled library evolution with score-based admission rather than unconstrained memory accumulation, and (iii) integrating multi-level credit assignment that exploits the same observation clustering used for retrieval.

\paragraph{Comparison with search-augmented RL.}
For search-augmented QA~\citep{kwiatkowski2019natural, joshi2017triviaqa, mallen2023trust, yang2018hotpotqa, ho2020constructing, trivedi2022musique, press2023measuring}, Search-R1~\citep{jin2025searchr1} trains agents to iteratively search and reason using RL with verifiable rewards. ZeroSearch~\citep{sun2025zerosearch} and StepSearch~\citep{zheng2025stepsearch} extend this paradigm with zero-shot and step-level search strategies, respectively. EvolveR~\citep{wu2025evolver} introduces evolving retrieval during training. These methods focus on improving the retrieval component of the agent, whereas \texttt{SLEA}-RL augments the agent with accumulated \emph{experiences}---distilled strategies and warnings from past episodes---providing complementary guidance that helps the agent reason more effectively over retrieved documents.

\FloatBarrier
\section{Broader Impact and Potential Risks}
\label{sec:appendix_broader_impact_risks}

\texttt{SLEA}-RL improves the ability of agents to learn from accumulated experience, which can benefit applications requiring multi-step reasoning and decision-making, such as virtual assistants, scientific discovery, and automation. However, the framework may also amplify biases or errors present in the collected trajectories, as these are distilled into reusable experiences and reused across tasks. In addition, experience reuse could lead to over-reliance on past patterns, reducing adaptability in novel or adversarial settings. The increased computational cost of step-level retrieval may further limit accessibility. Careful curation of experience data, monitoring of failure cases, and efficiency improvements are important to mitigate these risks.

\clearpage
\section{Main results tables}
\label{sec:appendix_tables}
\begin{table*}[!t]
    \caption{Performance on ALFWorld and WebShop. We report average success rate (\%) for ALFWorld subtasks and overall, and both average score and success rate (\%) for WebShop. $^*$: results from~\citet{feng2025group}. Best and second best in \textbf{bold} and \underline{underline}.}
    \label{tab:alfworld-webshop-full}
    \centering
    \setlength{\tabcolsep}{3.5pt}
    \renewcommand{\arraystretch}{0.92}
    \footnotesize
    \begin{tabular}{@{}ll ccccccc cc@{}}
    \toprule
    & & \multicolumn{7}{c}{\textbf{ALFWorld}} & \multicolumn{2}{c}{\textbf{WebShop}} \\
    \cmidrule(lr){3-9} \cmidrule(lr){10-11}
    \textbf{Type} & \textbf{Method} & Pick & Look & Clean & Heat & Cool & Pick2 & All & Score & Succ. \\
    \midrule
    \multicolumn{11}{@{}l}{\textit{Closed-Source LLMs}} \\
    & GPT-4o & 75.3 & 60.8 & 31.2 & 56.7 & 21.6 & 49.8 & 48.0 & 31.8 & 23.7 \\
    & Gemini-2.5-Pro & 92.8 & 63.3 & 62.1 & 69.0 & 26.6 & 58.7 & 60.3 & 42.5 & 35.9 \\
    \midrule
    \multicolumn{11}{@{}l}{\textit{Qwen2.5-7B-Instruct: Prompt / Memory Methods}} \\
    & ReAct$^*$ & 48.5 & 35.4 & 34.3 & 13.2 & 18.2 & 17.6 & 31.2 & 46.2 & 19.5 \\
    & Reflexion$^*$ & 62.0 & 41.6 & 44.9 & 30.9 & 36.3 & 23.8 & 42.7 & 58.1 & 28.8 \\
    & Mem0 & 54.0 & 55.0 & 26.9 & 36.4 & 20.8 & 7.69 & 33.6 & 23.9 & 2.00 \\
    & ExpeL & 21.0 & 67.0 & 55.0 & 52.0 & 71.0 & 6.00 & 46.3 & 30.9 & 11.2 \\
    & MemP & 54.3 & 38.5 & 48.1 & 56.2 & 32.0 & 16.7 & 41.4 & 25.3 & 6.40 \\
    & SimpleMem & 64.5 & 33.3 & 20.0 & 12.5 & 33.3 & 3.84 & 29.7 & 33.2 & 8.59 \\
    \midrule
    \multicolumn{11}{@{}l}{\textit{Qwen2.5-7B-Instruct: RL Methods}} \\
    & RLOO$^*$ & 87.6 & \underline{78.2} & 87.3 & 81.3 & 71.9 & 48.9 & 75.5 & 80.3 & 65.7 \\
    & GRPO$^*$ & 90.8 & 66.1 & 89.3 & 74.7 & 72.5 & 64.7 & 77.6 & 79.3 & 66.1 \\
    & GiGPO$^*$ & \underline{97.7} & 82.7 & \underline{98.8} & 83.7 & 89.3 & 79.2 & 90.8 & 84.4 & 72.8 \\
    & MemRL & 62.8 & 38.5 & 22.2 & 12.5 & 8.00 & 0.00 & 21.4 & 29.5 & 9.20 \\
    & Mem0+GRPO & 78.1 & 54.8 & 56.1 & 31.0 & 65.0 & 26.9 & 54.7 & 58.1 & 37.5 \\
    & SimpleMem+GRPO & 89.5 & 36.3 & 60.0 & 50.0 & 64.9 & 26.3 & 62.5 & 67.8 & 46.9 \\
    & R$^3$L & -- & -- & -- & -- & -- & -- & \underline{94.8} & -- & \underline{75.7} \\
    & SkillRL & 97.9 & 71.4 & 90.0 & \underline{90.0} & \underline{95.5} & \underline{87.5} & 89.9 & \underline{85.2} & 72.7 \\
    \rowcolor{ourscolor} & \texttt{SLEA}-RL & \textbf{98.5} & \textbf{85.1} & \textbf{99.0} & \textbf{93.8} & \textbf{96.4} & \textbf{90.4} & \textbf{93.5} & \textbf{87.6} & \textbf{76.3} \\
    \midrule
    \multicolumn{11}{@{}l}{\textit{Qwen2.5-1.5B-Instruct}} \\
    & GRPO$^*$ & 85.3 & 53.7 & 84.5 & 78.2 & 59.7 & 53.5 & 72.8 & 75.8 & 56.8 \\
    & GiGPO$^*$ & \underline{94.4} & 67.5 & \underline{94.8} & \underline{94.4} & 79.8 & \underline{76.4} & \underline{86.7} & 83.1 & 65.0 \\
    \rowcolor{ourscolor} & \texttt{SLEA}-RL & 92.6 & \textbf{80.0} & 96.2 & 94.7 & \textbf{95.0} & 76.5 & \textbf{87.5} & \textbf{88.7} & \textbf{75.4} \\
    \bottomrule
    \end{tabular}
    \vspace{-3mm}
    \end{table*}

    \vspace{-3mm}
    \begin{table*}[!t]
    \caption{Performance on search-augmented QA tasks. \texttt{SLEA}-RL is trained on NQ and HotpotQA. $^\dagger$: in-domain datasets. Best and second best among RL methods in \textbf{bold} and \underline{underline}.}
    \label{tab:search-qa-full}
    \centering
    \setlength{\tabcolsep}{4pt}
    \renewcommand{\arraystretch}{0.92}
    \footnotesize
    \begin{tabular}{@{}ll ccc c ccc c@{}}
    \toprule
    & & \multicolumn{3}{c}{\textbf{Single-Hop QA}} & & \multicolumn{3}{c}{\textbf{Multi-Hop QA}} & \\
    \cmidrule(lr){3-5} \cmidrule(lr){7-9}
    \textbf{Type} & \textbf{Method} & NQ$^\dagger$ & TriviaQA & PopQA & HotpotQA$^\dagger$ & 2Wiki & MuSiQue & Bamboogle & Avg. \\
    \midrule
    \multicolumn{10}{@{}l}{\textit{Qwen2.5-7B-Instruct: Prompting Methods}} \\
    & Qwen2.5 & 11.6 & 35.6 & 14.4 & 1.20 & 16.4 & 4.80 & 22.2 & 15.2 \\
    & CoT & 12.8 & 35.6 & 24.0 & 3.80 & 16.2 & 6.60 & 22.6 & 17.4 \\
    & RAG & 27.4 & 58.2 & 16.8 & 17.8 & 25.8 & 9.40 & 23.2 & 25.5 \\
    & Search-o1 & 19.4 & 40.6 & 30.4 & 11.4 & 17.0 & 8.60 & 27.0 & 22.1 \\
    & R1-Instruct & 21.0 & 44.9 & 19.2 & 17.1 & 20.8 & 6.00 & 27.5 & 22.4 \\
    & Search-R1 & 39.3 & 61.0 & 36.8 & 39.7 & 37.0 & 14.6 & 40.1 & 38.5 \\
    & ZeroSearch & 43.6 & 61.8 & 27.8 & 51.5 & 34.6 & 18.4 & 35.2 & 39.1 \\
    & EvolveR & 43.5 & 63.4 & 54.4 & 44.6 & 38.2 & 15.6 & 42.0 & 43.1 \\
    \midrule
    \multicolumn{10}{@{}l}{\textit{Qwen2.5-7B-Instruct}} \\
    & RLOO & 40.7 & 72.5 & 43.1 & 49.6 & 55.0 & 62.2 & 24.8 & 49.7 \\
    & PPO & 38.7 & 75.4 & 48.7 & 48.6 & 59.7 & 63.4 & 26.2 & 51.5 \\
    & GRPO & 40.3 & 77.0 & 49.6 & 48.9 & 57.7 & 65.1 & 25.0 & 51.9 \\
    & Reinforce++ & 34.3 & 67.5 & 44.3 & 45.9 & 54.5 & 61.2 & 23.7 & 47.3 \\
    & GSPO & 41.5 & 77.7 & 45.4 & 46.3 & 60.1 & 67.6 & 25.4 & 52.0 \\
    & GiGPO & 46.4 & 64.7 & 46.1 & 41.6 & 43.6 & 18.9 & \underline{68.9} & 47.2 \\
    & IGPO & \underline{46.7} & \underline{80.1} & 52.5 & \underline{57.2} & \underline{68.2} & 31.4 & \textbf{74.9} & \underline{58.7} \\
    & SkillRL & 45.9 & 63.3 & \textbf{73.8} & 45.9 & 43.2 & 20.2 & 40.3 & 47.1 \\
    \rowcolor{ourscolor} & \texttt{SLEA}-RL & \textbf{48.5} & \textbf{81.8} & \underline{55.2} & \textbf{59.8} & \textbf{70.5} & \textbf{77.2} & 33.2 & \textbf{60.9} \\
    \bottomrule
    \end{tabular}
    \vspace{-3mm}
    \end{table*}

\end{document}